\documentclass[journal]{IEEEtran}
\usepackage[utf8]{inputenc}
\usepackage[T1]{fontenc}
\usepackage{lmodern}
\usepackage{microtype}
\usepackage{amsmath,amssymb}
\usepackage{graphicx}
\usepackage{booktabs}
\usepackage{tabularx}
\usepackage{array}
\usepackage{multirow}
\usepackage{enumitem}
\usepackage{xcolor}
\usepackage{cite}
\usepackage{url}
\usepackage{hyperref}
\usepackage{balance}
\hypersetup{
pdftitle={Large language models transform organic synthesis from reaction prediction to automation},
pdfauthor={Kartar Kumar; Rajesh Kumar;  Nikesh Lagun}
}
\newcolumntype{Y}{>{\raggedright\arraybackslash}X}
\newcolumntype{P}[1]{>{\raggedright\arraybackslash}p{#1}}
\renewcommand{\arraystretch}{1.08}

\makeatletter
\def\abstract{\normalfont
    \if@twocolumn
      \@IEEEabskeysecsize\bfseries\textit{\abstractname:}\ \relax
    \else
      \bgroup\par\addvspace{0.5\baselineskip}\centering\vspace{-1.78ex}\@IEEEabskeysecsize\textbf{\abstractname}\par\addvspace{0.5\baselineskip}\egroup\quotation\@IEEEabskeysecsize
    \fi\@IEEEgobbleleadPARNLSP}
\def\endabstract{\relax\ifCLASSOPTIONconference\vspace{0ex}\else\vspace{1.34ex}\fi\par\if@twocolumn\else\endquotation\fi
    \normalfont\normalsize}
\def\IEEEkeywords{\normalfont
    \if@twocolumn
      \@IEEEabskeysecsize\bfseries\textit{\IEEEkeywordsname:}\ \relax
    \else
      \bgroup\par\addvspace{0.5\baselineskip}\centering\@IEEEabskeysecsize\textbf{\IEEEkeywordsname}\par\addvspace{0.5\baselineskip}\egroup\quotation\@IEEEabskeysecsize
    \fi\@IEEEgobbleleadPARNLSP}
\def\endIEEEkeywords{\relax\ifCLASSOPTIONtechnote\vspace{1.34ex}\else\vspace{0.67ex}\fi
    \par\if@twocolumn\else\endquotation\fi\normalfont\normalsize}
\makeatother
\title{Large language models transform organic synthesis from reaction prediction to automation}
\author{Kartar~Kumar, Rajesh~Kumar, and Nikesh~Lagun%
\thanks{K. Kumar is with the Department of Computer Science, Institute of Business Administration (IBA), Karachi, Pakistan.}%
\thanks{R. Kumar and N. Lagun are affiliated with the International Research Center for Complexity Sciences, Hangzhou International Innovation Institute, Beihang University, Hangzhou 311115, China, and the School of Computer Science and Engineering, University of Electronic Science and Technology of China, Chengdu 611731, China.}%
}

\begin{document}
\maketitle

\begin{abstract}
Large language models (LLMs) are beginning to reshape how organic-synthesis workflows are represented, queried, planned, and connected to experimental automation. Assessing their contribution is not straightforward because reaction-specific transformers, chemistry-adapted LLMs, tool-using agents, optimizers, and autonomous laboratories are often discussed under the same broad terminology despite operating at different levels of the synthesis workflow. This Review traces the progression from reaction prediction and retrosynthetic planning to reaction development, literature-to-protocol translation, and robotic execution, with representative numerical claims verified against primary sources available through 4 September 2026. We examine where language models provide genuine added value, where established specialist methods remain the principal source of performance, and how retrieval, uncertainty-aware optimization, deterministic chemistry tools, robotics, sensing, and human oversight alter system behavior. The evidence shows a clear transition from isolated language-model demonstrations toward modular scientific systems that combine broad language priors with specialized computation and experimentally grounded feedback. Emerging work on multimodal reaction understanding, standardized tool interoperability, provenance-aware execution, and programmable laboratory infrastructure extends this transition further. Despite this progress, reliable deployment continues to depend on generalization beyond historical reaction corpora, calibrated uncertainty, transparent component attribution, safety controls, reproducible system state, and prospective experimental validation.
\end{abstract}

\begin{IEEEkeywords}
large language models, reasoning models, agentic AI, organic synthesis, retrosynthesis, multimodal reaction understanding, uncertainty-aware optimization, scientific agents, self-driving laboratories, cheminformatics, provenance.
\end{IEEEkeywords}

\section{Introduction}
The integration of large language models (LLMs) with organic synthesis has expanded the interface between computational chemistry and experimental automation. This development, however, enters a field with a long history of computer-assisted reasoning. Since the early work of Corey and Wipke on computer-aided synthesis planning \cite{corey1969}, synthetic route design has progressed through expert systems, data-driven reaction prediction, graph and sequence models, heuristic search, and optimization \cite{szymkuc2016casp,dealmeida2019ai,long2025retrosynthesisreview}. Contemporary LLMs therefore represent a new layer within an established computational ecosystem rather than the origin of computer-assisted synthesis.

The pre-LLM lineage is particularly important for evaluating novelty. Sequence-to-sequence models, graph-based predictors, neural-symbolic planners, and attention-based reaction transformers had already established template-free reaction prediction and data-driven retrosynthesis before conversational LLMs became common \cite{schwaller2018translation,segler2018planning,schwaller2019molecular}. The Molecular Transformer showed that reaction outcomes could be learned without hand-coded reaction rules \cite{schwaller2019molecular}, while Chemformer extended pretraining across several molecular and reaction tasks \cite{irwin2022chemformer}. These advances established many of the predictive capabilities that are now incorporated into broader agentic systems.

What changed after 2023 was the range of functions that could be coordinated through language. General-purpose and chemistry-adapted LLMs brought natural-language interaction, in-context learning, retrieval, code generation, and tool calling into chemical workflows \cite{jablonka2024predictive,bran2024chemcrow,ramos2025review}. In parallel, language models were connected to synthesis planners, scientific software, automated instruments, and closed-loop experimentation \cite{yoshikawa2023robotics,boiko2023coscientist,ruan2024llmrdf}. By 2025 and 2026, the landscape included chemistry-focused platforms such as SynAsk \cite{zhang2025synask}, chemistry-adapted models such as Chemma \cite{zhang2025chemma}, reaction foundation models such as ReactionT5 \cite{sagawa2025reactiont5}, multi-agent robotic systems \cite{song2025chemagents}, and hybrid workflows for retrosynthesis, optimization, procedure compilation, and self-correcting laboratory control \cite{sathyanarayana2026deepretro,ramos2026boicl,pagel2026chemputation,panapitiya2026autolabs}. The resulting progression is not simply from smaller to larger models; it is from isolated prediction toward coordinated scientific systems.

This progression also creates an attribution problem. A reaction transformer trained on SMILES is not equivalent to a conversational LLM, and a route produced by an LLM that calls a retrosynthesis engine reflects the combined system rather than the language model alone. The same issue appears in autonomous experimentation. When Bayesian optimization selects experiments, deterministic software controls hardware, and analytical instruments provide feedback, experimental performance emerges from the interaction of these components. Similar care is required for numerical comparisons because results obtained with different data sets, splits, prompts, model versions, decoding settings, and experimental endpoints cannot be interpreted as a common leaderboard.

Recent reviews have examined LLMs and autonomous agents across chemistry \cite{ramos2025review,alampara2026gpm,macknight2025rethinking}, and a 2026 \emph{Chemical Reviews} article analyzes organic chemistry as a driver of AI methodology, including reaction prediction, chemical reasoning, and agentic AI \cite{chawla2026organicai}. Other reviews emphasize molecular synthesis with LLMs \cite{xu2026molecularsynthesis}, LLM-guided reaction development \cite{takebe2026reactiondevelopment}, or self-driving laboratories \cite{canty2026sdl}. These perspectives establish the breadth of the field but leave a specific need for an execution-oriented synthesis: how evidence changes as a system moves from molecular and reaction representations to prediction, planning, optimization, protocol translation, and physical action.

\subsection{Scope and Organization}
This Review addresses that need by organizing the literature along the organic-synthesis execution pathway. The analysis separates reaction-specific models, chemistry-adapted LLMs, tool-using agents, optimizers, and embodied laboratory systems, allowing performance to be attributed to the component or combination of components that actually produces it. Representative quantitative claims are traced to primary publications, and numerical comparisons are retained only when the underlying task, metric, and experimental context can be reconstructed. The literature is updated through 4 September 2026 and includes emerging developments in multimodal reaction understanding, uncertainty-aware optimization, self-correcting agents, Model Context Protocol (MCP) interoperability, and programmable laboratory infrastructure. A parallel objective is to establish an evidence hierarchy that distinguishes textual plausibility and benchmark recovery from chemically grounded validation, prospective experimentation, and independent replication.

The organization follows the increasing complexity of the synthesis workflow. Section II defines the evidence base and terminology used throughout the Review. Section III examines the progression from chemical sequence models to chemistry-adapted LLMs, followed in Section IV by reaction prediction, retrosynthesis, benchmark interpretation, and reaction-data limitations. Sections V and VI consider strategic synthesis planning and reaction development, including established non-LLM optimization baselines. Section VII addresses literature intelligence and protocol digitization, while Section VIII examines robotics and autonomous experimentation. Sections IX and X focus on evidence quality, reproducibility, reliability, interpretability, safety, and sustainability. Section XI discusses the late-2026 frontier and emerging directions toward 2027. Section XII develops reporting priorities and testable research propositions, and Section XIII summarizes the conditions required for reliable deployment.

\begin{table*}[t]
\caption{Scope of this Review relative to recent high-level reviews. The comparison identifies differences in emphasis and does not assert priority.}
\label{tab:priorreviews}
\centering
\footnotesize
\renewcommand{\arraystretch}{0.92}
\begin{tabularx}{\textwidth}{P{3.0cm}P{1.1cm}YYYY}
\toprule
\textbf{Review} & \textbf{Year} & \textbf{Primary scope} & \textbf{LLM/agent coverage} & \textbf{Physical automation} & \textbf{Distinctive emphasis} \\
\midrule
Ramos \emph{et al.} \cite{ramos2025review} & 2025 & Chemistry broadly & Extensive & Yes & General history, applications, tools, and scientific agents. \\
Alampara \emph{et al.} \cite{alampara2026gpm} & 2026 & General-purpose chemical models & Extensive & Selected examples & Model building principles, chemical data, evaluation, safety and ethics. \\
Chawla \emph{et al.} \cite{chawla2026organicai} & 2026 & AI innovation motivated by organic chemistry & Extensive & Yes & Data sparsity, reaction prediction, chemical reasoning, agentic AI. \\
Xu \emph{et al.} \cite{xu2026molecularsynthesis} & 2026 & Molecular synthesis with LLMs & Central & Yes & Knowledge, data structuring, prediction, synthesis planning and execution. \\
Canty and Abolhasani \cite{canty2026sdl} & 2026 & Self-driving laboratories & Trustworthy AI agents as one component & Central & Scalability, generalizability, infrastructure, provenance. \\
\textbf{This review} & 2026 & \textbf{Organic-synthesis execution stack} & \textbf{Central} & \textbf{Central} & \textbf{Source-verified evidence ladder; model, agent, and robot decomposition; reaction-to-execution continuity.} \\
\bottomrule
\end{tabularx}
\end{table*}

\section{Scope, Evidence Base, and Terminology}
\subsection{Evidence Base and Verification}
The Review uses a structured critical approach designed for a rapidly evolving literature in which model names, benchmark versions, preprints, software platforms, and experimental systems frequently overlap. The evidence base combines foundational work that established reaction modeling, retrosynthesis, optimization, and laboratory automation with peer-reviewed primary studies published through 4 September 2026. Benchmark papers are included when they materially affect interpretation of chemical-model performance, and recent high-level reviews are used to position the scope of the present synthesis. Preprints are limited to frontier topics for which a peer-reviewed equivalent was not available at the literature cutoff and are identified explicitly. Official infrastructure announcements are discussed only as indicators of changing research infrastructure, not as evidence of chemical performance.

Primary publications provide the basis for quantitative statements wherever possible. Numerical values are retained only when the underlying task, metric, data set, and evaluation context can be reconstructed from the source. Claims lacking traceable provenance, including unsupported percentages, ambiguous system names, or industrial anecdotes without a primary record, are omitted. This source-level discipline is important in a fast-moving field because numerical precision can otherwise obscure substantial differences in experimental design.

Evidence collection proceeded from broad mapping to source verification. Recent reviews and perspectives were first used to identify the main synthesis-relevant themes: molecular and reaction representation, foundation models, forward prediction, retrosynthesis, condition and yield modeling, scientific agents, laboratory automation, benchmarking, and safety. Searches then targeted primary studies using combinations of synthesis, reaction prediction, retrosynthesis, optimization, chemical agents, autonomous laboratories, and individual system names. Backward and forward citation tracing recovered foundational studies that predate the current LLM wave. Priority was given to work that introduced a synthesis-relevant model or system, defined an influential benchmark, reported prospective experimental validation, or changed the interpretation of autonomy, interoperability, or reproducibility.

A final verification pass focused on nomenclature and causal attribution. System names were checked against the corresponding primary publications, benchmark names were disambiguated, and model-level claims were separated from platform-level outcomes. Numerical results were removed when the denominator, task definition, or experimental setting could not be established. The resulting corpus is intentionally broad in conceptual coverage but conservative in quantitative synthesis, which allows historical development and emerging architectures to be discussed without treating heterogeneous evidence as directly comparable.

\subsection{Operational Definitions}
Table~\ref{tab:definitions} defines the units used throughout the Review. The categories are based on function rather than on loosely defined labels such as ``chemistry-specific'' or ``task-oriented,'' which can overlap within a single system.

\begin{table*}[t]
\caption{Operational distinctions used in this review.}
\label{tab:definitions}
\centering
\small
\renewcommand{\arraystretch}{1.0}
\begin{tabularx}{\textwidth}{P{2.7cm}YYP{3.2cm}}
\toprule
\textbf{System level} & \textbf{Operational definition} & \textbf{Typical contribution} & \textbf{Representative examples} \\
\midrule
Reaction-specific transformer / chemical foundation model & Model trained primarily on molecular or reaction representations for defined predictive or generative chemical tasks. It need not support open-ended dialogue. & Forward reaction prediction, single-step retrosynthesis, yield prediction, learned reaction embeddings. & Molecular Transformer \cite{schwaller2019molecular}; Chemformer \cite{irwin2022chemformer}; ReactionT5 \cite{sagawa2025reactiont5}. \\
Chemistry-adapted LLM & A generative language model adapted by fine-tuning, instruction data, retrieval, or domain resources to support chemistry-related natural-language or structured tasks. & Chemical QA, literature assistance, reaction prediction, route discussion, code generation. & ChemLLM (preprint) \cite{zhang2024chemllm}; SynAsk \cite{zhang2025synask}; Chemma \cite{zhang2025chemma}. \\
Tool-augmented agent & An LLM embedded in a software loop that can select external tools, retrieve information, execute code, call chemistry software, or iteratively revise a plan. & Planning, verification, synthesis-tool use, data analysis, experiment selection. & ChemCrow \cite{bran2024chemcrow}; Coscientist \cite{boiko2023coscientist}; DeepRetro \cite{sathyanarayana2026deepretro}. \\
Embodied or autonomous experimental system & A system that connects decision-making software to physical laboratory equipment. The LLM may be only one component among planners, optimizers, robotics controllers, sensors, and safety interlocks. & Procedure translation, robotic execution, closed-loop optimization, experiment scheduling, characterization. & CLAIRify/chemistry robotics \cite{yoshikawa2023robotics}; LLM-RDF \cite{ruan2024llmrdf}; ORGANA \cite{darvish2025organa}; ChemAgents \cite{song2025chemagents}; AutoLabs \cite{panapitiya2026autolabs}. \\
\bottomrule
\end{tabularx}
\end{table*}

The term \emph{reasoning} is used here in an operational sense. It denotes observable performance on tasks that require multi-step inference, constraint satisfaction, strategy specification, or tool-mediated planning, and does not imply human-like cognition. \emph{Mechanistic understanding} is reserved for stronger evidence: a system must generate or discriminate chemically plausible mechanistic hypotheses under controlled evaluation. Attention weights, fluent rationales, or a correct final product are not sufficient by themselves.

\subsection{System-Level Architecture of LLM-Enabled Synthesis}
\begin{figure*}[t]
\centering
\includegraphics[width=0.96\textwidth]{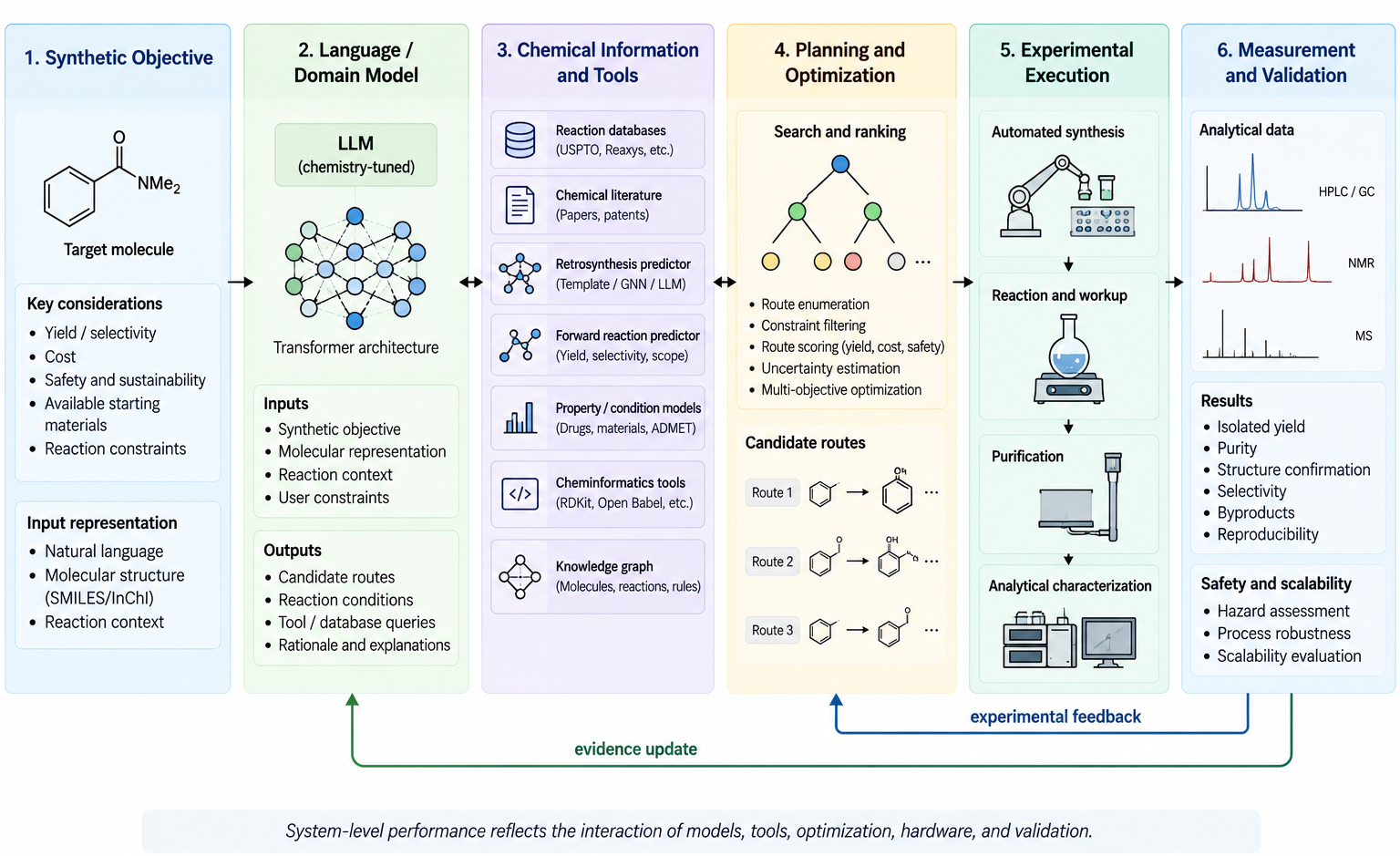}
\caption{LLM-enabled organic-synthesis execution stack, from objective specification and chemistry-adapted language modeling to specialist tools, route optimization, physical execution, and analytical validation. The feedback paths emphasize that reliable synthesis emerges from the complete system rather than from the language model in isolation.}
\label{fig:stack}
\end{figure*}

These definitions lead naturally to the system-level decomposition shown in Figure~\ref{fig:stack}. The language model can interpret goals, retrieve information, or propose actions, but reliable synthesis usually depends on additional layers for chemical knowledge, deterministic tools, optimization, device control, sensing, analytical validation, and safety. Experimental performance consequently belongs to the complete architecture unless component-level ablations establish a narrower source of the observed gain.

\begin{figure*}[t]
\centering
\includegraphics[width=0.96\textwidth]{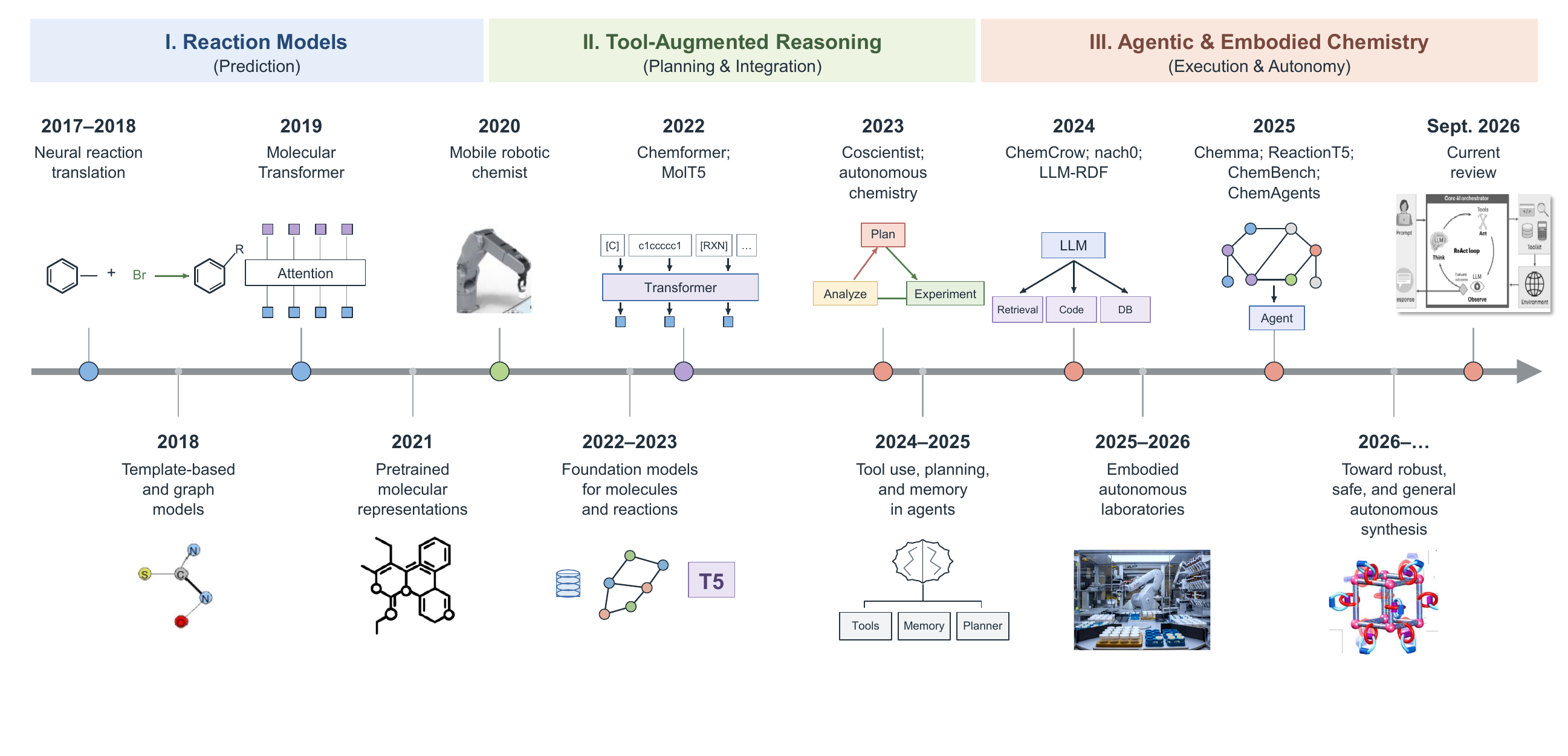}
\caption{Selected milestones from reaction-specific language models to tool-augmented and embodied chemical AI. The chronology emphasizes that contemporary LLM-enabled synthesis builds on earlier advances in reaction transformers, computer-assisted synthesis planning, digital chemistry, and autonomous laboratories \cite{schwaller2018translation,schwaller2019molecular,burger2020mobile,irwin2022chemformer,edwards2022molt5,boiko2023coscientist,bran2024chemcrow,ruan2024llmrdf,zhang2025chemma,sagawa2025reactiont5,song2025chemagents,sathyanarayana2026deepretro,panapitiya2026autolabs}.}
\label{fig:timeline}
\end{figure*}

\section{From Chemical Sequence Models to Chemistry-Adapted LLMs}
\subsection{Pre-LLM Foundations}
The transition from rule-based systems to learned synthesis models occurred through several distinct methodological steps. Data-driven retrosynthesis already included molecular-similarity retrieval \cite{coley2017similarity} and recurrent sequence-to-sequence prediction \cite{liu2017retroseq}. Graph-based approaches learned reaction-center or template applicability directly from molecular structure, including graph-convolutional reactivity prediction \cite{coley2019reactivity} and the Conditional Graph Logic Network \cite{dai2019cgl}. Attention-based reaction models also recovered useful structure from reaction strings; RXNMapper, for example, inferred atom correspondences without supervised atom-mapping labels \cite{schwaller2021grammar}. Template-free prediction, attention over reaction strings, and learned chemical representations were therefore established before instruction-tuned conversational models became common.

At the representation level, chemical reactions can be represented as strings, graphs, sets, or combinations of these modalities. SMILES remains a widely used linear representation \cite{weininger1988smiles}, while SELFIES was designed to guarantee syntactic molecular validity under its grammar \cite{krenn2020selfies}. Treating reaction prediction as sequence translation enabled neural models to learn transformations directly from data \cite{schwaller2018translation}. The Molecular Transformer then demonstrated strong reaction prediction using attention without manually encoded reaction templates \cite{schwaller2019molecular}. Neural-symbolic systems simultaneously advanced retrosynthetic search \cite{segler2018planning}.

These developments establish two baselines for the present discussion. Treating chemistry as a language-like sequence problem predates contemporary LLMs, and self-attention is an architectural operation rather than direct evidence of chemical reasoning \cite{vaswani2017attention}. Whether a model encodes chemically meaningful structure must be established through task performance, controlled perturbations, representation analysis, or prospective validation.

Building on this progression, Chemformer showed that pretraining a transformer on molecular/reaction data can support multiple downstream sequence and discriminative tasks \cite{irwin2022chemformer}. ReactionT5 extends this line of work using Open Reaction Database pretraining. Its authors report 97.5\% accuracy for product prediction, 71.0\% for retrosynthesis, and $R^2=0.947$ for yield prediction under their reported benchmark settings \cite{sagawa2025reactiont5}. Those figures establish the value of reaction-focused pretraining, while the architecture and interface place ReactionT5 more naturally among reaction foundation models than among open-ended conversational LLMs.

\subsection{General-Purpose LLMs and Domain Adaptation}
In contrast to specialist reaction models, general-purpose LLMs can be adapted to chemistry by fine-tuning or by expressing structured chemical tasks as natural-language questions. Jablonka \emph{et al.} demonstrated that a general LLM could be fine-tuned for multiple predictive chemistry tasks, with especially competitive behavior in low-data regimes \cite{jablonka2024predictive}. The result is conceptually important because it shows that broad language priors can become useful chemical predictors without constructing a bespoke architecture for every target property.

Among the early domain-adapted efforts, ChemLLM, released as an arXiv preprint in 2024, represents an early effort to build an open chemistry-focused conversational LLM using instruction-formatted chemical data \cite{zhang2024chemllm}. Because the work remains a preprint, it is best treated as evidence of model-development direction rather than as a peer-reviewed benchmark standard. It also introduced a benchmark called ChemBench. This resource is distinct from the independently developed, peer-reviewed ChemBench framework published in \emph{Nature Chemistry} in 2025 \cite{mirza2025chembench}.

By comparison, SynAsk illustrates a different design: a domain-specific organic-chemistry platform that combines LLM fine-tuning with a knowledge base and chemistry tools for molecular retrieval, reaction performance prediction, retrosynthesis, and literature access \cite{zhang2025synask}. Calling SynAsk simply a ``model'' obscures the importance of this external-resource integration. Its unit of evaluation is better understood as a platform.

More tightly integrated with synthesis tasks, Chemma represents a chemistry-adapted LLM. The model was fine-tuned using 1.28 million reaction question--answer pairs and evaluated across synthesis-related tasks \cite{zhang2025chemma}. The primary publication reports 72.2\% Top-1 accuracy for single-step retrosynthesis on USPTO-50K and integrates Chemma with Bayesian optimization and active learning. In a human-AI reaction-development case study, a suitable ligand and solvent for an unreported Suzuki--Miyaura coupling were identified within 15 experimental runs, yielding 67\% isolated product \cite{zhang2025chemma}. These are substantially stronger and better-defined claims than vague statements that a model ``understands chemistry.''

The 2025 literature also shows that scale and training strategy can strengthen reaction-specific models without resolving the boundary between an LLM and a specialist synthesis model. RSGPT pre-trained a generative transformer on more than ten billion reaction tokens and reported strong performance across reaction tasks \cite{zheng2025rsgpt}. Its contribution lies in reaction-scale pretraining rather than open-ended scientific dialogue. This distinction changes the appropriate baseline. Specialist reaction models are evaluated primarily against specialist predictors, whereas conversational models additionally require evidence for instruction following, tool use, and behavior outside a fixed input-output task.

A second distinction concerns retrospective recovery versus prospective capability. Strong performance on USPTO-derived benchmarks can reflect effective learning of common historical transformations without establishing robustness to new reaction classes, altered conditions, or laboratory execution. Prospective experiments therefore occupy a different evidentiary level from held-out benchmark recovery, even when both are reported as synthesis performance.

Terminology is also shifting as the 2025 and 2026 literature distinguishes conventional instruction-tuned LLMs from \emph{large reasoning models} (LRMs) or reasoning-oriented inference configurations. Sharlin \emph{et al.}, for example, evaluated both categories on NMR structure-elucidation problems that require integration of multiple spectral constraints \cite{sharlin2026nmr}. The label is operational rather than epistemic. Additional inference-time computation or a longer reasoning trace does not by itself demonstrate chemically faithful reasoning; the same requirements for calibration, counterfactual testing, and prospective validation still apply.

\subsection{Chemical Language, Instruction Tuning, and Multimodality}
Beyond reaction-specific models, the wider molecular-language literature provides additional architectural context. ChemBERTa explored large-scale self-supervised transformer pretraining on molecular strings \cite{chithrananda2020chemberta}; Mol-BERT used bidirectional transformer representations for molecular-property prediction \cite{li2021molbert}; and MoLFormer scaled molecular-language pretraining to very large unlabeled molecular corpora \cite{ross2022molformer}. Text--molecule integration then broadened through MolXPT \cite{liu2023molxpt}, MoleculeSTM \cite{liu2023moleculestm}, MolCA \cite{liu2023molca}, and BioT5 \cite{pei2023biot5}. MolPROP provides a complementary example in which learned language and graph representations are fused for molecular-property prediction \cite{rollins2024molprop}. These models are not all synthesis planners, but they define the representation and multimodal design space from which synthesis-oriented foundation models increasingly draw.

Between specialist reaction transformers and general conversational models lies a broad family of intermediate architectures. MolT5 jointly pre-trained on molecular strings and natural language, establishing a shared molecular-text representation space \cite{edwards2022molt5}. nach0 extended this idea to multimodal chemical and natural-language tasks \cite{livne2024nach0}. Collectively, these models show that architecture, training corpus, interface, and evaluation task are distinct descriptors and lose information when compressed into a single ``chemical LLM'' category.

Another route to domain specialization is illustrated by ChemDFM. It was trained on 34 billion tokens from chemical literature and subsequently fine-tuned with 2.7 million instructions, with the goal of retaining conversational flexibility while increasing chemical-domain competence \cite{chemDFM2025}. This scale makes ChemDFM relevant to synthesis-support tasks that require literature knowledge or free-form dialogue, but it also reinforces the need to distinguish broad chemical proficiency from direct evidence on experimentally consequential synthesis tasks. ChemLLM provides an open chemistry-focused conversational model, although its 2024 version remains a preprint. Its internal ChemBench benchmark is separate from the later peer-reviewed benchmark of the same name \cite{zhang2024chemllm,mirza2025chembench}.

Multimodality is especially relevant because chemical practice is not text-only. Molecular depictions, reaction schemes, spectra, chromatograms, tables, and instrument interfaces carry information that may be lost in a SMILES-only workflow. ChemVLM introduced a chemistry-oriented multimodal model for textual and visual chemical information \cite{li2025chemvlm}, and ChemMLLM later pursued a unified interface across text, SMILES, and molecular images \cite{tan2026chemMLLM}. These systems do not constitute evidence of end-to-end autonomous synthesis, but they address an important upstream requirement for scientific agents that must integrate symbolic molecular representations with the visual and analytical modalities encountered in real laboratories.

Representation remains a scientific design choice rather than a formatting detail. SMILES offers a compact linearization but is sensitive to tokenization and string non-uniqueness; SELFIES prioritizes syntactic robustness; graph representations encode connectivity directly; and reaction-specific languages can expose transformations that remain implicit in reactant and product strings. ReactSeq, for example, represents reactions as sequences of molecular editing operations \cite{zhang2025reactseq}. A larger model cannot always compensate for a representation that obscures the variable of interest. Explicit reaction-center or edit information may simplify reaction prediction, while structured procedure languages can provide stronger execution guarantees than free-form prose in laboratory automation.

\subsection{Attributing Contributions to Models and Systems}
A recurring methodological difficulty is the boundary between transformer-based chemical models and LLMs. Neither parameter count nor tokenization provides a sufficient boundary. We use a functional distinction: reaction-specific models are optimized primarily for defined chemical mappings, whereas LLMs are expected to support broader instruction following, natural-language interaction, or task generalization. Foundation models such as ReactionT5 and Chemformer occupy an intermediate position because pretraining supports multiple chemical tasks while the interface remains largely structured and task-specific \cite{irwin2022chemformer,sagawa2025reactiont5}. The appropriate evaluation standard follows from this function. Reaction predictors are judged on validity, accuracy, calibration, and generalization; agents additionally require evidence for tool selection, planning, recovery, and interaction with external state.

\begin{figure*}[t]
\centering
\includegraphics[width=0.94\textwidth]{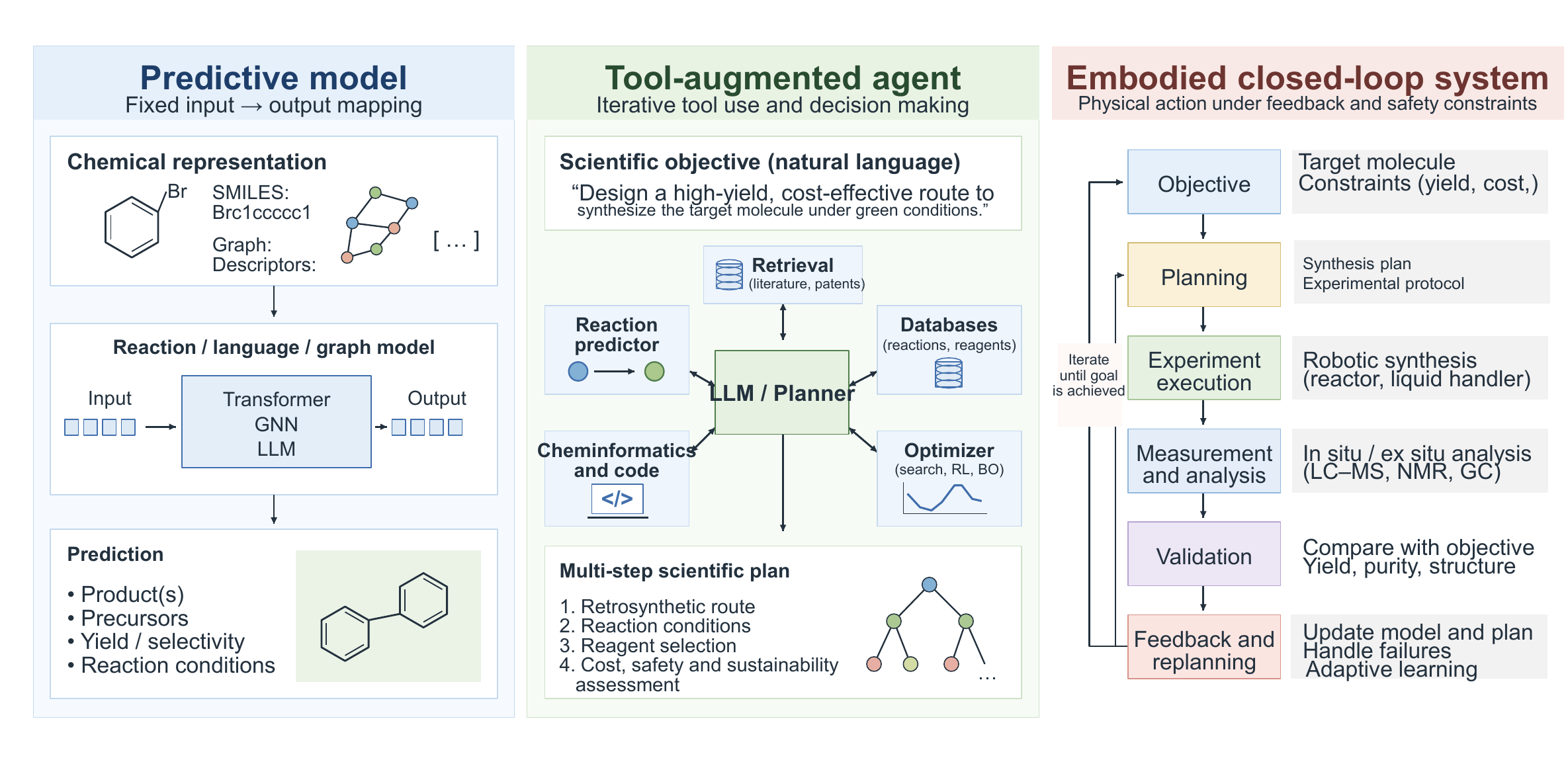}
\caption{Progression from predictive models to tool-augmented agents and embodied closed-loop synthesis systems. The comparison separates decision horizon, evidence source, physical authority, typical outputs, and principal limitations, thereby clarifying why model-level accuracy alone is insufficient to characterize agentic or autonomous chemical systems.}
\label{fig:modelagent}
\end{figure*}

This distinction also constrains novelty claims. An agent that calls a state-of-the-art retrosynthesis engine contributes orchestration, interface design, verification, or closed-loop integration, not the retrosynthetic prediction generated by the underlying engine. Conversely, gains in yield prediction or low-data transfer are informative only relative to strong specialist baselines, not generic chat models alone. Clear attribution identifies the component responsible for an observed gain and makes subsequent improvements scientifically interpretable. Figure~\ref{fig:modelagent} summarizes this progression from fixed predictive mappings to tool-mediated planning and finally to embodied closed-loop experimentation.

\section{Reaction Prediction and Retrosynthesis}
Reaction prediction and retrosynthesis span a heterogeneous methodological lineage. Template-free route planning has been coupled to transformer prediction \cite{lin2020autosynroute,schwaller2020hypergraph}; neural-guided search has developed through methods such as Retro* \cite{chen2020retrostar}; and single-step diversity or plausibility has been addressed with tied two-way transformers \cite{kim2021tied}, graph-edit models such as MEGAN \cite{sacha2021megan}, and Graph2Edits \cite{zhong2023graph2edits}. Search remains an active problem, including MCTS-enhanced A* variants \cite{zhao2024mcts}. Industrial experience with AiZynthFinder 4.0 further shows that practical route quality depends on search configuration, stock definitions, reaction policies, and workflow integration rather than on a single-step Top-$k$ score alone \cite{saigiridharan2024aizynth4}. Stereochemical disconnections remain challenging even for mature CASP systems \cite{wiest2024stereo}, while recent dual-task language-model formulations explore joint reaction and retrosynthesis prediction \cite{lin2025chemdual}.

\subsection{Distinguishing Reaction-Prediction Tasks}
The phrase ``reaction prediction'' covers several distinct problems. \emph{Forward prediction} maps reactants, and sometimes reagents or conditions, to products. \emph{Single-step retrosynthesis} maps a target product to plausible precursor sets. \emph{Multi-step retrosynthesis} adds route search, termination criteria, purchasability constraints, and global route ranking. \emph{Condition prediction} estimates catalysts, solvents, reagents, temperature, time, or related variables, while \emph{yield prediction} is a regression problem whose reliability depends strongly on how the underlying experiments were generated and measured.

These tasks have different output spaces and different failure costs. Exact string match is informative for standardized single-step benchmarks but does not measure multi-step synthetic usefulness. A chemically reasonable alternative disconnection may also be counted as incorrect when it differs from the route recorded in a patent. Benchmark accuracy consequently represents one level of evidence, not a surrogate for laboratory success.

\subsection{Verified Representative Quantitative Evidence}
Table~\ref{tab:quant} summarizes representative primary results for which the task and evaluation context can be stated clearly. The values are presented as study-specific evidence and are not ranked across rows.

\begin{table*}[t]
\caption{Representative quantitative results verified against primary publications. Values are study-specific; the rows do not constitute a common leaderboard.}
\label{tab:quant}
\centering
\footnotesize
\renewcommand{\arraystretch}{0.9}
\begin{tabularx}{\textwidth}{P{2.4cm}P{1.2cm}P{3.0cm}YP{4.0cm}}
\toprule
\textbf{System} & \textbf{Year} & \textbf{Task / data} & \textbf{Reported result} & \textbf{Interpretive constraint} \\
\midrule
ReactionT5 \cite{sagawa2025reactiont5} & 2025 & Product prediction, retrosynthesis, yield prediction & 97.5\% product-prediction accuracy; 71.0\% retrosynthesis accuracy; yield $R^2=0.947$ & Reaction-specific foundation model; metrics derive from the authors' benchmark settings and are not directly comparable with conversational LLM evaluations. \\
Chemma \cite{zhang2025chemma} & 2025 & USPTO-50K single-step retrosynthesis & 72.2\% Top-1 accuracy & Result is specific to the reported training/evaluation protocol; cross-paper comparison requires split and preprocessing equivalence. \\
Chemma + active learning \cite{zhang2025chemma} & 2025 & Experimental Suzuki--Miyaura reaction exploration & Ligand and solvent identified within 15 runs; 67\% isolated yield & Demonstrates a human-AI experimental loop, not autonomous unrestricted synthesis. \\
Electrochemical C--H oxidation workflow \cite{zheng2025electrochem} & 2025 & Reactivity prediction and condition optimization & ML reactivity classifier reported $>$90\% accuracy; high-yield conditions identified for eight drug-like substances/intermediates & Core prediction was performed by ML; LLMs supported literature mining, code generation, and function calling. \\
DeepRetro \cite{sathyanarayana2026deepretro} & 2026 & Hybrid multi-step retrosynthesis & Strong route-solving performance reported on multi-step benchmark sets and complex case studies & Hybrid system integrates LLMs, conventional retrosynthesis engines, validity checks, and expert feedback; performance reflects the combined architecture. \\
BO-ICL \cite{ramos2026boicl} & 2026 & Bayesian optimization using frozen LLMs as in-context surrogates & In live RWGS catalyst experiments, near-equilibrium CO yield was approached within 6 and 10 iterations for candidate pools of 3,700 and 360,000, respectively & Catalysis/materials optimization rather than classical organic retrosynthesis; illustrates LLM-enabled optimization without weight updates. \\
\bottomrule
\end{tabularx}
\end{table*}

\begin{figure*}[t]
\centering
\includegraphics[width=0.86\textwidth]{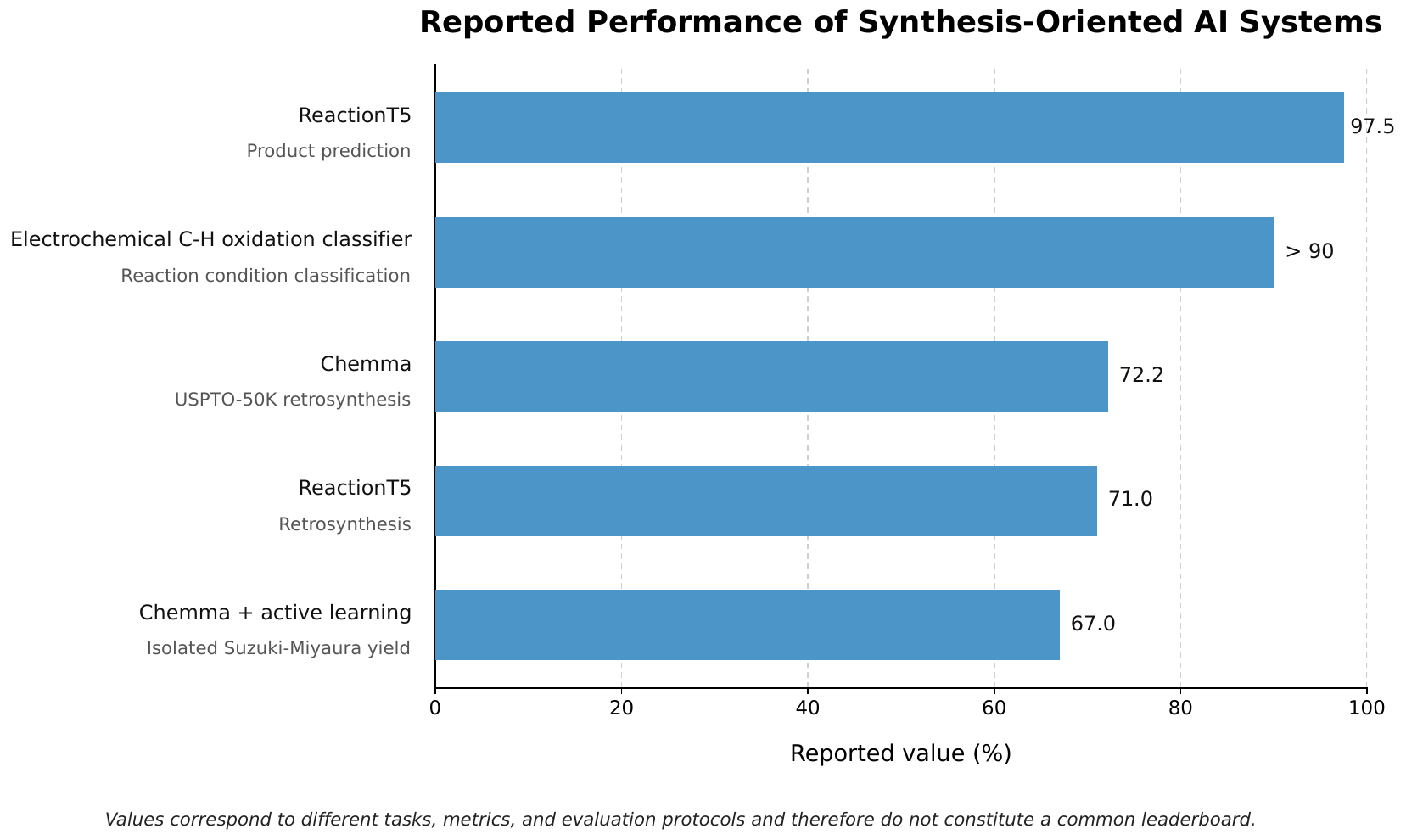}
\caption{Representative quantitative results retained after primary-source verification \cite{sagawa2025reactiont5,zhang2025chemma,zheng2025electrochem}. The bars share only a reported percentage scale; the underlying tasks, data sets, and protocols differ, so the figure summarizes evidence without implying a common ranking.}
\label{fig:quantresults}
\end{figure*}

\subsection{Limits of Cross-Study Comparison}
Patent-derived reaction data sets, including widely used USPTO corpora \cite{lowe2017uspto}, contain historical and curation biases that complicate cross-study comparison. Reported performance can change with atom mapping, deduplication, stereochemical handling, reagent separation, canonicalization, and train-test leakage. General-purpose LLMs introduce further variables, including prompt wording, system instructions, model version, tool access, temperature, and sampling budget. A difference in reported accuracy is meaningful only when these sources of variation are controlled or disclosed.

Meaningful cross-study comparison requires the exact model or checkpoint, data-set version and split, task formulation, input representation, prompt or instruction template, tool access, decoding settings, sample size, metric definition, repeated-run variability where feasible, and the source of each value. When those details are unavailable, qualitative synthesis is more defensible than a unified performance hierarchy.

\subsection{ChemBench and the Scope of General Chemistry Evaluation}
ChemBench illustrates why benchmark scope must be stated explicitly. The peer-reviewed benchmark contains 2,788 chemistry question-answer pairs that test knowledge, reasoning, calculation, and intuition across chemistry topics \cite{mirza2025chembench}. It is useful for evaluating broad chemical competence and tool-augmented systems, but it is not an organic reaction data set analogous to USPTO-50K and does not substitute for direct evaluation of products, precursors, yields, or synthetic routes.

The ChemBench results also caution against interpreting aggregate scores as general chemical reliability. Leading models performed strongly on the overall benchmark, yet weaknesses remained on basic tasks, domain-specific knowledge, preference judgments, and confidence calibration \cite{mirza2025chembench}. Broad chemical competence and reliable synthesis behavior are related but distinct evaluation targets.

\subsection{Multimodal Reaction Understanding from Scientific Literature}
Authentic scientific documents introduce a further challenge beyond reaction strings. Models must align chemical structures, reaction schemes, tables, captions, and surrounding prose before they can recover a coherent experimental account. RxnBench directly probes this setting using 1,525 questions derived from 305 reaction schemes together with full-document tasks drawn from 108 papers \cite{li2026rxnbench}. No evaluated model exceeded 50\% on the full-document task, indicating that document-level multimodal understanding remains a substantial bottleneck upstream of reliable literature-to-protocol automation.

\subsection{Reaction Representations and Chemical Validity}
Benchmark accuracy also hides the distinction between string correctness and chemical validity. A forward model can produce a syntactically valid string that violates atom balance or reaction plausibility, while a retrosynthesis model can miss an equivalent precursor representation or propose a plausible disconnection absent from the recorded route. The Molecular Transformer partly addressed reliability by associating predictions with calibrated uncertainty \cite{schwaller2019molecular}. Later methods such as Retroformer emphasize reactive regions through combined sequence and graph information \cite{wan2022retroformer}, whereas ReactSeq represents transformations as molecular editing operations \cite{zhang2025reactseq}. These approaches show that representation influences both the type of errors a model makes and the form of validation that is appropriate.

For a review article, the implication is straightforward: Top-1 accuracy on USPTO-50K is interpretable only within a specified preprocessing and split convention. Even when the data-set name is shared, reaction-class labels, atom mapping, canonicalization, stereochemical treatment, and duplicate removal can alter the task. Multi-step solve rates add further dependence on purchasable-stock definitions and route-scoring rules. Expert assessment and prospective execution become increasingly important as evaluation moves from string recovery toward synthetic strategy.

\subsection{Reaction Data: Patents, ORD, and Experimental Heterogeneity}
Data-set selection is part of the scientific design. Comparative work shows that template specificity, coverage, and route-finding performance depend strongly on the source corpus \cite{thakkar2020datasets}, and learned synthetic-complexity measures inherit the reaction distributions on which they are trained \cite{coley2018scscore}. Interpretation of benchmark results therefore depends on corpus provenance, preprocessing, atom mapping, stock definitions, and the treatment of negative or failed experiments.

In particular, the USPTO corpus enabled a large fraction of modern reaction modelling, but patent data are a historical record of what was disclosed rather than a balanced sampling of chemical reactivity \cite{lowe2017uspto}. Successful reactions are overrepresented; negative experiments are sparse; procedure text may be incomplete; reaction classes are uneven; and the same transformation may appear multiple times with different levels of detail. These biases become especially important when a model is asked to extrapolate to new substrate classes or optimize conditions rather than merely reproduce common transformations.

The Open Reaction Database (ORD) provides a complementary source of structured reaction data. Its open schema supports conventional bench chemistry, high-throughput experimentation, and flow workflows \cite{kearnes2021ord}, and it has subsequently supported reaction-focused pretraining, including ReactionT5 \cite{sagawa2025reactiont5}. Open structured data remain smaller and more heterogeneous than proprietary industrial archives. High-throughput experimentation offers dense local landscapes with controlled variation and often includes low-yield outcomes, making it particularly useful for yield prediction and optimization. Examples include nanomole-scale automated screening \cite{perera2018nanomole} and HTE modeling of Buchwald-Hartwig reaction performance \cite{ahneman2018cn}. Because such data sets often cover narrow reaction families, high interpolation performance provides limited evidence for broad generalization. Evaluation is stronger when it includes random, scaffold, reaction-family, temporal, and, where possible, external-laboratory splits.

\subsection{From Benchmark Performance to Prospective Utility}
The preceding limitations define an evidence hierarchy from benchmark recovery to chemical utility. At the lowest level, a model reproduces strings from a held-out corpus. Stronger tests verify atom balance, molecular validity, stereochemistry, and reaction-center consistency. Expert assessment can then evaluate alternative routes under explicit constraints. Prospective evidence begins when a model influences an experiment that was not part of the benchmark corpus, and confidence increases further when successful execution is supported by analytical characterization and independent replication. This hierarchy is used below to compare agentic and autonomous systems.

\section{From Retrosynthetic Output to Strategy-Aware Planning}
Single-step disconnections capture only a fraction of the decisions involved in synthesis planning. Route design also depends on protecting-group strategy, convergence, functional-group compatibility, route length, starting-material availability, safety, scalability, and the timing of key bond formations. Tool-augmented LLMs are attractive in part because many of these preferences can be expressed in natural language and combined with algorithmic search.

ChemCrow, for example, is an instructive case. It couples GPT-4 to 18 expert-designed chemistry tools rather than relying on the language model as a closed-book oracle \cite{bran2024chemcrow}. The system planned and executed syntheses of DEET and three organocatalysts and supported chromophore discovery. The scientific lesson is not that GPT-4 alone became a synthetic chemist; it is that an LLM can act as an orchestration layer over specialized tools when the system constrains tool selection, information retrieval, and action sequencing.

More recently, a 2026 study in \emph{Matter} directly investigated natural-language strategic specifications for synthesis planning and reaction-mechanism search \cite{bran2026reasoning}. The reported framework achieved 71\% alignment with independent expert chemists on synthesis-strategy judgments and used LLM-guided search to identify plausible mechanisms. This is stronger evidence for strategy-aware behavior than attention visualization because the construct is externally evaluated against expert judgments. Even here, ``reasoning'' remains an operational term: alignment with chemists and successful search are measurable outcomes, whereas claims of human-like mechanistic cognition extend beyond the available evidence.

DeepRetro provides a complementary example of hybrid planning \cite{sathyanarayana2026deepretro}. The framework combines LLM generation with conventional retrosynthetic engines, validity checks, iterative refinement, and expert feedback. Its architecture illustrates a recurring design principle: generative flexibility is most useful when it is bounded by chemistry-specific search and verification rather than treated as a complete synthetic plan.

Mechanistic exploration is beginning to converge with this planning literature. Chen \emph{et al.} reported an LLM-guided workflow for reaction-pathway exploration in 2025 that couples language-model guidance with computational chemistry \cite{chen2025pathway}. Mechanistic hypotheses are converted into candidates for explicit computational evaluation rather than accepted directly from the model. The language model therefore guides the search space, while physics-based or quantum-chemical calculations determine whether proposed pathways are energetically and structurally plausible.

\subsection{Synthesis Planning as Constrained Search}
From an algorithmic perspective, CASP is a search problem over a branching space of disconnections. Neural prediction and symbolic search were combined successfully before modern LLM agents, including the neural-symbolic planning framework of Segler \emph{et al.} \cite{segler2018planning} and open-source planners such as AiZynthFinder \cite{genheden2020aizynthfinder}. A language model that proposes a plausible retrosynthetic step is therefore not equivalent to a planning system that must recursively expand nodes, avoid cycles, rank complete routes, enforce stock constraints, and balance route quality against search cost.

Within this search formulation, LLMs are potentially valuable because strategic preferences are difficult to encode in a fixed scalar objective. A chemist may prefer convergence over linearity, late-stage diversification over early commitment, robust transformations over maximum atom economy, or a route that avoids a hazardous reagent even at the cost of one extra step. Natural language provides a flexible interface for such constraints. The 2026 strategy-aware framework in \emph{Matter} is important precisely because it uses the LLM as an evaluator within search rather than as an unconstrained route generator \cite{bran2026reasoning}. This architecture preserves the combinatorial precision of algorithmic search while allowing human strategy to enter through language.

A complementary hybrid approach is provided by DeepRetro, which iterates between LLM proposals, conventional CASP components, validity checking, and expert feedback \cite{sathyanarayana2026deepretro}. The resulting system is best understood as hybrid intelligence, with its advantage arising from coordination between generative and symbolic components. Such systems also make failure analysis more tractable. If a route fails because the LLM suggested an unstable intermediate, that is a different error from a search heuristic failing to explore a viable branch or an inventory system incorrectly rejecting an accessible precursor.

Route-level evaluation extends beyond Top-$k$ step accuracy. Relevant endpoints include solve rate under a fixed search budget, route diversity, convergence, protection and deprotection burden, stereochemical fidelity, realistic starting materials, hazard burden, estimated cost, and expert or experimental preference. No single endpoint captures synthetic quality completely, but a multidimensional route profile is more informative than treating the patent route as the only acceptable answer.

\section{Reaction Development and Condition Optimization}
\subsection{Non-LLM Baselines for Reaction Optimization}
Reaction optimization was already an algorithmic discipline before current LLM agents. Neural models have recommended reaction conditions from large reaction databases \cite{gao2018conditions}, and deep reinforcement learning has been used for iterative reaction optimization \cite{zhou2017drl}. Contemporary reviews emphasize complementary roles for design of experiments, mechanistic and kinetic models, Bayesian optimization, high-throughput experimentation, and self-optimization \cite{taylor2023optimization}. The relevant baselines for LLM-enabled optimization are strong purpose-built methods, not only unaided language models or manual trial-and-error.

Current interest in LLM-guided experiment selection builds on a mature literature in data-efficient chemical optimization. Phoenics introduced Bayesian optimization for expensive chemistry objectives \cite{hase2018phoenics}, and Gryffin extended related ideas to categorical and mixed-variable design spaces relevant to catalyst, ligand, and process selection \cite{hase2021gryffin}. Shields \emph{et al.} subsequently benchmarked Bayesian reaction optimization against human decision-making and applied it prospectively to synthetic reaction-development problems \cite{shields2021bo}. These studies establish demanding baselines for determining whether language-derived representations or priors improve experiment selection.

The central comparison is therefore not LLM versus random search. It is whether language-derived priors, in-context learning, or semantically rich representations improve on Gaussian processes, chemically informed descriptors, categorical Bayesian optimization, design of experiments, or other established sequential strategies. LLMs may be advantageous when the search space contains complex categorical variables, sparse literature knowledge, or changing natural-language constraints. Conventional Bayesian methods may remain preferable when the design space is well parameterized and calibrated uncertainty is the primary concern. The most informative studies compare these approaches under the same experimental budget.

\subsection{LLMs within Sequential Optimization Loops}
Reaction development extends beyond route proposal to condition selection, interpretation of analytical data, hypothesis refinement, and selection of the next experiment. Language models can contribute through literature mining, code generation, structured extraction, surrogate modeling, or experiment proposal. The scientific contribution, however, depends on identifying the decision mechanism responsible for experiment selection.

Within this setting, Chemma provides one of the clearer synthesis-focused demonstrations because the language model was integrated with Bayesian optimization and active learning, and the paper reports an experimental reaction-development case rather than only offline benchmark scores \cite{zhang2025chemma}. Still, the loop is a composite system: model predictions, optimizer behavior, experimental execution, and human collaboration all contribute to the final result.

Likewise, Zheng \emph{et al.} combined high-throughput electrochemistry, machine learning, and LLMs to explore electrochemical C--H oxidation \cite{zheng2025electrochem}. LLMs were used to mine literature and generate code/function calls, while conventional ML models performed core reactivity prediction with reported accuracy above 90\%. The study optimized conditions for eight drug-like substrates or intermediates and assembled 1,071 reaction-condition/yield data points. In this workflow, the LLM contributes experimental informatics and workflow acceleration; the electrochemical reactivity function itself is learned by the specialist ML component.

BO-ICL provides a particularly clear formalization of a language model within sequential optimization. The 2026 method extends in-context learning to regression with uncertainty estimates and uses frozen LLMs as surrogate models inside Bayesian optimization \cite{ramos2026boicl}. It matched or exceeded Gaussian-process baselines on several benchmark problems and was also tested in live reverse water-gas shift experiments, where catalysts approaching equilibrium CO yield were identified within 6 and 10 iterations from candidate pools of 3,700 and 360,000, respectively. Although the chemistry is oriented toward catalysis and materials, the methodological lesson transfers directly to reaction development: language-derived representations can be embedded in an explicit exploration-exploitation framework with uncertainty. BO-ICL is best understood as a Bayesian optimization architecture in which the LLM serves a defined statistical role.

GOLLuM, published on 28 August 2026, moves further in this direction by training language-model representations with Bayesian objectives so that experimental outcomes and uncertainty reshape the learned space \cite{rankovic2026gollum}. Across 23 tasks spanning organic synthesis, materials science, process chemistry, and molecular design, the study reports first-ranked average performance among the compared methods, matches traditional Bayesian optimization with more than 40\% fewer experiments, and improves the discovery rate of high-performing Buchwald-Hartwig reactions relative to descriptor and LLM baselines. The important conceptual change is the treatment of uncertainty as part of the optimization objective rather than an afterthought to fluent generation.

Complementing these model-centered approaches, a JCIM study by Mottafegh and Ahn moves human tacit knowledge into the same uncertainty-aware framework \cite{mottafegh2026hitl}. Their human-in-the-loop approach uses an LLM to translate natural-language expert hypotheses into structured prior mean functions and constraints for Bayesian optimization, then applies adaptive weight credibility detection (AWCD) to test whether those priors remain consistent with incoming experimental data. When evidence contradicts an expert-supplied prior, the mechanism can suppress it and recover data-driven behavior. The authors validate the framework on a continuous Ugi multicomponent-reaction problem and a discrete P3HT/CNT optimization setting. For synthesis, the important architectural lesson is that the LLM need not be the optimizer: it can serve as a semantic interface that converts qualitative chemical knowledge into probabilistic structure while a calibrated sequential decision layer determines what experiment to run next.

For organic synthesis, this uncertainty-aware direction may be more consequential than continued scaling of conversational models. Reaction development is a low-data sequential decision problem in which the cost of one experiment can exceed the cost of many model inferences. An optimizer must quantify not only its prediction but also its uncertainty and the information expected from the next experiment. Language-derived representations are valuable when they transfer useful chemical priors across sparse tasks; they are less compelling when a conventional Gaussian process with chemically meaningful descriptors is equally sample efficient. Direct comparisons under matched experimental budgets are needed to establish the added value of the language model.

\subsection{Autonomy Beyond Language Models}
Modern self-driving laboratories also show that closed-loop autonomy does not require an LLM. RoboChem-Flex combines modular hardware, Python-based real-time control, and Bayesian optimization and supports both autonomous and human-in-the-loop modes across several reaction classes \cite{pilon2026robochemflex}. Its relevance is methodological: adaptive experiment selection and laboratory control can be achieved through optimization and engineering alone. Within this setting, the contribution of an LLM is measured by gains in accessibility, transfer, semantic interpretation, recovery, or scientific productivity relative to a strong non-LLM autonomous baseline.

\section{Literature Intelligence, Data Extraction, and Protocol Digitization}
Literature-to-laboratory automation begins with the conversion of experimental prose into structured operations. Transformer-based extraction of synthesis actions showed that procedures can be mapped to explicit operation sequences \cite{vaucher2020actions}. This distinction between text generation and procedure representation is fundamental: executable operations can be constrained by an ontology, quantities and dependencies can be checked, and the resulting protocol can be tested against actual hardware capabilities.

Much of synthetic knowledge nevertheless remains distributed across prose, tables, schemes, supplementary files, and inconsistent procedure formats. This creates an information bottleneck before reaction prediction or robotic execution can begin. LLMs are well suited to converting unstructured language into structured representations, but chemistry places stricter demands on this transformation than generic information extraction because stoichiometry, order of operations, atmosphere, work-up, and analytical context can determine experimental success.

ReactionSeek addresses this upstream bottleneck through LLM-powered literature mining for organic synthesis \cite{li2026reactionseek}. The system is designed to extract reaction information at scale and support downstream knowledge discovery rather than merely summarize papers. High-quality extraction can improve model training, reaction-condition databases, substrate-scope analysis, and evidence retrieval for synthesis agents. In this sense, the value of an LLM may lie as much in recovering structured knowledge from historical literature as in generating new reaction hypotheses.

A more difficult downstream problem is converting a literature procedure into an executable laboratory protocol. Prose written for human chemists contains tacit assumptions: the order and rate of additions may be implied; atmospheric conditions may appear several sentences earlier; concentrations may require calculation; work-up language may be qualitative; and hardware-specific constraints are typically unstated. An LLM can propose a structured translation, but executable automation requires a formal intermediate representation and hardware-aware validation.

For example, Pagel \emph{et al.} demonstrated a 2026 workflow in which an LLM-based research agent extracts synthetic procedures and analytical data, translates procedures into XDL, simulates hardware-specific execution, and then executes selected literature syntheses on chemputation platforms \cite{pagel2026chemputation}. The study reports six realistic literature-derived syntheses executed on two robotic platforms. This provides an unusually clear evidence chain: unstructured literature $\rightarrow$ structured procedure $\rightarrow$ simulation $\rightarrow$ physical execution. It also illustrates why a formal procedure language can be more important than conversational fluency. The LLM handles ambiguity and translation, whereas XDL and the robotic platform constrain execution.

Literature-to-protocol translation therefore forms a distinct layer of the synthesis stack and requires its own evaluation. Relevant measures include extraction precision and recall, stoichiometric consistency, preservation of operation order, recovery of conditions and work-up details, executable-code validity, hardware transferability, and physical reproduction. Success at document extraction does not guarantee success at laboratory execution, and a single ``agent accuracy'' score would obscure the source of downstream failure.

\section{LLM-Enabled Robotics and Autonomous Experimentation}
The automation lineage in chemistry substantially predates contemporary LLM agents. Active-learning and robotic experimentation were used to explore crystallization spaces \cite{duros2017robots}; machine-learning-controlled synthesis robots searched reaction space \cite{granda2018robot}; and reactionware encoded multistep pharmaceutical synthesis in digitally specified hardware \cite{kitson2018reactionware}. Reconfigurable systems enabled automated reaction optimization \cite{bedard2018reconfigurable}, modular robotic platforms linked chemical programming languages to organic-synthesis execution \cite{steiner2019robotic}, and ChemOS provided an orchestration layer for autonomous experimentation \cite{roch2018chemos}. These developments form the experimental baseline for assessing LLM-enabled robotics.

Integration of synthesis planning with automation also predates current agentic systems. An AI-informed robotic flow platform connected route planning to automated execution \cite{coley2019roboticflow}. Digital chemistry then progressed from a universal literature-to-execution architecture \cite{mehr2020universal} to validated executable reaction databases in the ChemPU \cite{rohrbach2022chempu} and a portable universal synthesis platform \cite{manzano2022portable}. The self-driving-laboratory literature accordingly defines autonomy through closed-loop integration of prediction, automation, measurement, and feedback rather than through the presence of an LLM \cite{abolhasani2023sdl}.

\subsection{From Natural-Language Instructions to Laboratory Execution}
Physical execution introduces constraints absent from text-only benchmarks, including collisions, liquid handling, device state, timing, contamination, calibration, uncertainty, and safety. Yoshikawa \emph{et al.} combined LLM-based translation of natural-language instructions with a domain-specific representation, program verification, constrained task and motion planning, perception, and robotic execution \cite{yoshikawa2023robotics}. Reliability arose from the complete architecture rather than from unconstrained natural-language generation.

Similarly, Coscientist combined GPT-4 with web/documentation access, code execution, and automated laboratory hardware \cite{boiko2023coscientist}. The system demonstrated several chemistry tasks, including autonomous planning and execution and closed-loop optimization of palladium-catalyzed cross-coupling reactions. The work helped establish that general-purpose LLMs can coordinate heterogeneous scientific tools, but it also underscores the need to report which steps are performed by the LLM versus external software and instrumentation.

Building on this idea, LLM-RDF broadened the architecture into an end-to-end reaction-development framework comprising six specialized agents: Literature Scouter, Experiment Designer, Hardware Executor, Spectrum Analyzer, Separation Instructor, and Result Interpreter \cite{ruan2024llmrdf}. The platform was demonstrated on copper/TEMPO aerobic alcohol oxidation and tested on additional $S_\mathrm{N}$Ar, photoredox C--C coupling, and heterogeneous photoelectrochemical tasks. This architecture is better characterized as an agentic laboratory workflow than as a single LLM.

By comparison, ORGANA is an assistive robotic system in which LLMs provide a natural-language interface for extracting goals, resolving ambiguity, and producing logs, while task/motion planning and perception support execution \cite{darvish2025organa}. It demonstrated solubility, pH, recrystallization, and electrochemistry experiments, including a 19-step electrochemistry plan. In a user study, ORGANA reduced frustration and physical demand by more than 50\% and saved users an average of 80.3\% of their time \cite{darvish2025organa}. These results belong to ORGANA under its published name and are not evidence for a separate ``ChemBot'' system.

Moving further toward distributed autonomy, ChemAgents uses a hierarchical multi-agent architecture built around an on-board Llama-3.1-70B model, with a Task Manager coordinating Literature Reader, Experiment Designer, Computation Performer, and Robot Operator agents \cite{song2025chemagents}. The system was demonstrated on six experimental tasks spanning synthesis, characterization, parameter exploration, and functional-material discovery. Again, the relevant scientific unit is the multi-agent robotic system, not the base language model in isolation.

More recently, two 2026 developments sharpen the reliability question. Pagel, Jirasek, and Cronin developed an LLM-augmented chemputation workflow that extracts synthetic procedures and analytical information from literature, translates them into a machine-readable chemical language, checks hardware-specific execution, and runs procedures on robotic platforms \cite{pagel2026chemputation}. AutoLabs instead focuses on the quality of language-to-protocol translation. It uses a self-correcting multi-agent architecture to convert natural-language instructions into hardware-ready liquid-handler files; across five benchmark experiments and 20 agent configurations, the study reports that stronger reasoning models reduced quantitative preparation errors by more than 85\%, and multi-agent/self-correction configurations approached expert-authored procedures with F1 scores above 0.89 on challenging multi-plate syntheses \cite{panapitiya2026autolabs}.

\begin{table*}[t]
\caption{Verified examples of LLM-enabled physical or near-physical chemical automation.}
\label{tab:robotics}
\centering
\footnotesize
\renewcommand{\arraystretch}{0.92}
\begin{tabularx}{\textwidth}{P{2.4cm}P{1.1cm}P{3.1cm}YY}
\toprule
\textbf{System} & \textbf{Year} & \textbf{LLM role} & \textbf{Non-LLM components that matter} & \textbf{Physical evidence reported} \\
\midrule
Chemistry robotics / CLAIRify \cite{yoshikawa2023robotics} & 2023 & Natural language $\rightarrow$ structured plan & Program verification, task/motion planning, perception, robotic skills & Real-robot execution of laboratory tasks including solubility and recrystallization. \\
Coscientist \cite{boiko2023coscientist} & 2023 & Planning, tool selection, code/information use & Web/documentation tools, code, laboratory automation & Multiple autonomous chemistry tasks and closed-loop reaction optimization. \\
LLM-RDF \cite{ruan2024llmrdf} & 2024 & Six specialized agents across reaction-development stages & Automated equipment, spectral analysis, separation workflow, user interface & End-to-end development of aerobic alcohol oxidation plus three additional reaction scenarios. \\
ORGANA \cite{darvish2025organa} & 2025 & Goal interpretation, disambiguation, natural-language reporting & Task/motion planning, perception, scheduling, robotics & Solubility, pH, recrystallization and electrochemistry; 19-step electrochemistry plan and user-study time savings. \\
ChemAgents \cite{song2025chemagents} & 2025 & Hierarchical multi-agent coordination & Literature/protocol/model resources and automated lab & Six experimental tasks from synthesis/characterization to screening and materials optimization. \\
Chemputation + LLM agent \cite{pagel2026chemputation} & 2026 & Literature procedure extraction and translation & Machine-readable chemical language, simulation/verification, robotic chemputation & Literature procedures translated, checked and executed on robotic platforms. \\
AutoLabs \cite{panapitiya2026autolabs} & 2026 & Dialogue, decomposition, stoichiometric reasoning, self-correction & Tool-assisted calculations and high-throughput liquid-handler interface & Five benchmark experiments; 20 architecture configurations; hardware-ready outputs and granular reliability evaluation. \\
\bottomrule
\end{tabularx}
\end{table*}

The historical automation baseline remains important when interpreting recent agentic systems. Burger \emph{et al.} reported a mobile robotic chemist in 2020 that autonomously conducted 688 photocatalyst experiments over eight days and identified formulations six times more active than the initial candidates using batched Bayesian search \cite{burger2020mobile}. Physical autonomy, adaptive experiment selection, and long-duration operation were demonstrated before the current use of LLMs. Digital chemistry and chemputation likewise established machine-readable procedures and programmable synthesis as core enabling technologies.

This baseline changes the appropriate role assigned to the language model in robotics. LLMs are most useful where laboratories contain semantic variability: interpreting a scientist's intent, reading documentation, translating prose into commands, selecting software tools, or coordinating heterogeneous instruments. They are less suitable as direct low-level controllers for safety-critical continuous variables. Coscientist follows this division by using language models for planning, documentation search, code, and orchestration while programmable laboratory interfaces execute physical actions \cite{boiko2023coscientist}. LLM-RDF similarly distributes reaction development across literature scouting, experiment design, hardware execution, spectral analysis, separation guidance, and result interpretation \cite{ruan2024llmrdf}.

From a human-interaction perspective, ORGANA emphasizes human accessibility and robotic assistance, allowing natural-language interaction for experimental planning while integrating manipulation, scheduling, perception, and characterization \cite{darvish2025organa}. ChemAgents moves further toward multi-agent autonomous research on demand by connecting AI planning with robotic execution \cite{song2025chemagents}. AutoLabs in 2026 makes self-correction itself an object of study, using a multi-agent architecture to translate natural-language instructions into liquid-handler protocols and evaluating reliability at a more granular level \cite{panapitiya2026autolabs}. Together these systems show a shift from the question ``Can an LLM call a robot?'' toward ``Can the overall agent architecture detect and recover from errors before they propagate into physical experiments?''

At the same time, two 2026 non-LLM-centric platforms provide useful counterpoints. RoboChem-Flex is a low-cost, modular self-driving laboratory that integrates device control with Bayesian optimization and supports fully closed-loop or human-in-the-loop reaction optimization \cite{pilon2026robochemflex}. Flex-Cat couples automated pressurized batch experimentation with hierarchical Bayesian optimization and executed 680 experiments in homogeneous catalysis across three multi-objective campaigns \cite{bennett2026flexcat}. These systems show that autonomy can advance through hardware modularity, optimization, and digital twins even when LLMs are peripheral or absent. The relevant comparison is whether adding an LLM improves robustness, accessibility, transfer, or scientific productivity relative to a strong non-LLM autonomous baseline.

Evaluation of self-driving laboratories is also becoming more explicit. The ADePT framework proposes adaptability and learning, dexterity, perception, and task complexity as four dimensions for assessing autonomous laboratory robotics \cite{salazar2026adept}. Canty and Abolhasani identify scalability, generalizability, and provenance-complete experimentation as requirements for the next phase of self-driving laboratories \cite{canty2026sdl}. These criteria are directly relevant to LLM-enabled systems because language competence represents only one dimension of embodied performance. A platform may generate a convincing procedure yet remain vulnerable to tube placement, sensor drift, clogged lines, instrument downtime, or reagent variability.

\subsection{Operational Levels of Laboratory Autonomy}
Tool augmentation itself also requires controlled evaluation. CACTUS combines language models with cheminformatics tools \cite{mcnaughton2024cactus}, while ChemAgent uses self-updating memories to support multi-step chemical reasoning \cite{tang2025chemagent}. Broad evaluation of chemistry agents indicates that tools do not improve every task uniformly: specialist tools help when their capabilities match the question, whereas other tasks remain limited by the underlying model \cite{yu2025tooling}. Modular agents are also appearing in adjacent drug-discovery workflows \cite{ock2026agent}. The term ``agentic'' therefore describes an architecture rather than an assumed improvement in quality.

To avoid treating every natural-language interface as an autonomous laboratory, we use the following operational levels:
\begin{enumerate}[leftmargin=*,label=\textbf{L\arabic*:}]
\item \textbf{Advisory}: the system proposes information or procedures; a human selects and executes all actions.
\item \textbf{Tool-augmented computational}: the system can call software, databases, or code but has no physical actuation.
\item \textbf{Supervised physical execution}: machine-generated procedures control equipment, but a human authorizes key steps or resolves exceptions.
\item \textbf{Closed-loop bounded autonomy}: the system selects and executes experiments within a predefined safe domain using automated measurements and feedback.
\item \textbf{Extended autonomous research}: the system can reformulate plans, recover from failures, and pursue multi-stage objectives with minimal human intervention, while still operating under explicit safety, resource, and access controls.
\end{enumerate}
An LLM-generated plan alone does not place a system at the highest level of autonomy. Autonomy must be demonstrated through action authority, feedback, recovery, and constraints.

\begin{figure*}[t]
\centering
\includegraphics[width=0.96\textwidth]{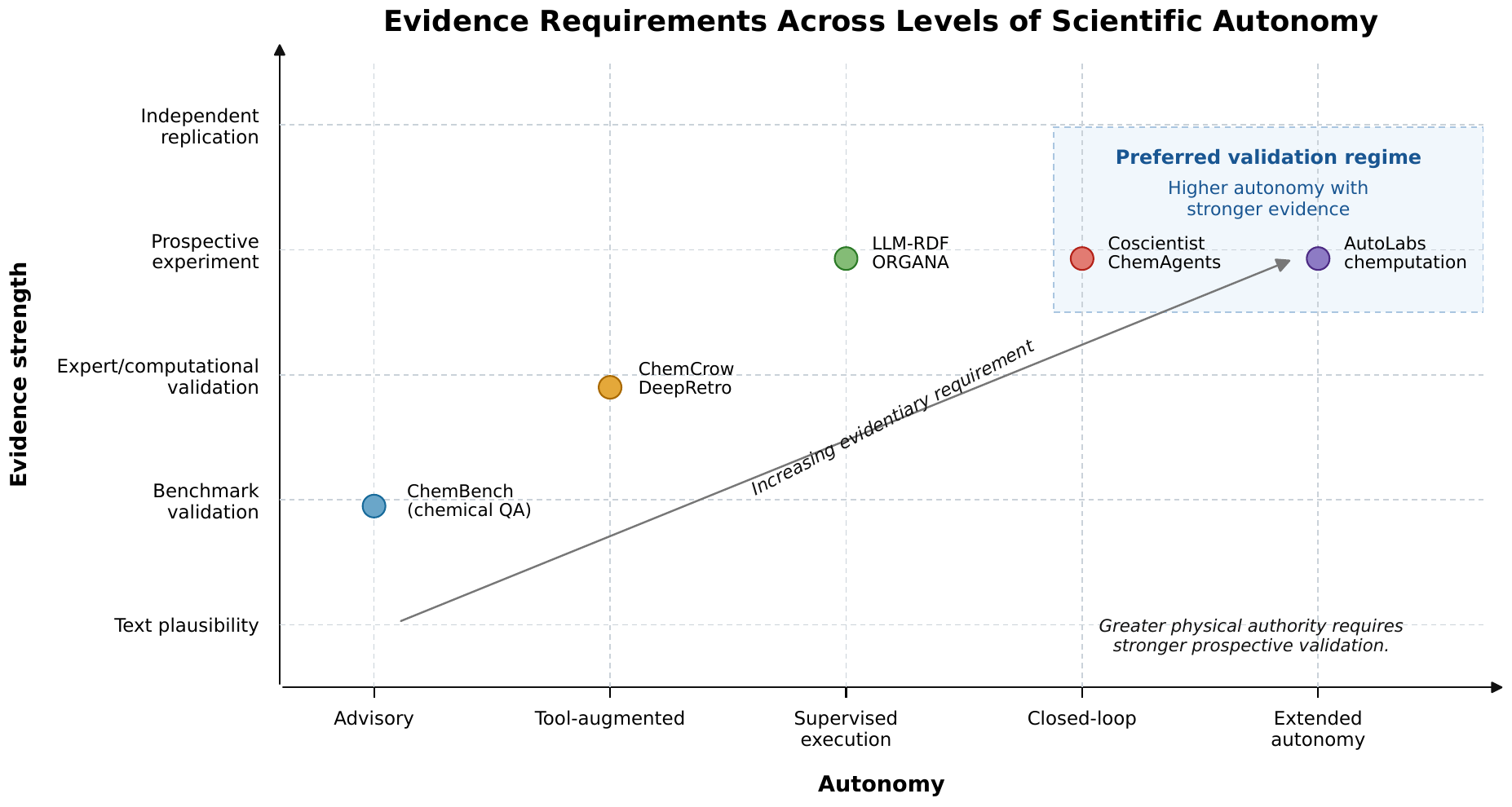}
\caption{Conceptual evidence-autonomy matrix. Autonomy and evidentiary strength are independent dimensions. Increasing authority over physical experiments raises the corresponding requirement for prospective validation, failure reporting, and independent replication. Placements are illustrative rather than numerical rankings.}
\label{fig:evidenceautonomy}
\end{figure*}

\section{Evaluation, Reproducibility, and Evidence}
\subsection{Evidence Levels for Synthesis Claims}
\begin{figure*}[t]
\centering
\includegraphics[width=0.76\textwidth]{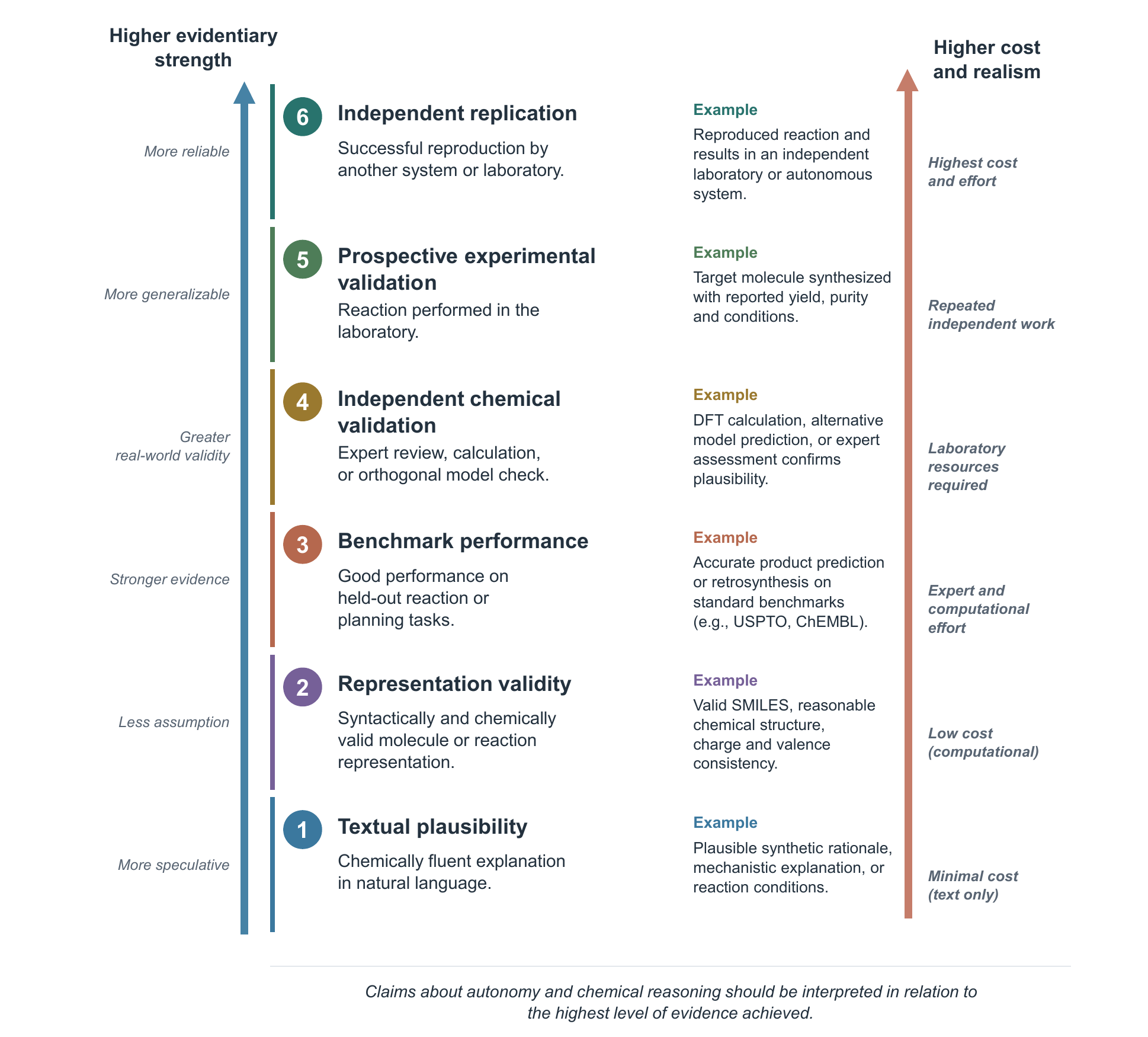}
\caption{Evidence hierarchy for synthesis-oriented AI claims. Textual plausibility and representation validity are useful preliminary checks, but stronger claims require benchmark validation, orthogonal chemical assessment, prospective experimentation, and ultimately independent replication. The hierarchy also reflects the increasing experimental cost and real-world realism associated with stronger evidence.}
\label{fig:evidenceladder}
\end{figure*}

Discussions of LLM-enabled chemistry frequently combine evidence that belongs to different scientific levels. We therefore distinguish six levels, ranging from textual plausibility to independent replication:
\begin{enumerate}[leftmargin=*]
\item \textbf{Textual plausibility}: an answer sounds chemically reasonable.
\item \textbf{Representation validity}: generated SMILES/reaction encodings are syntactically valid and chemically parseable.
\item \textbf{Benchmark agreement}: outputs match a labeled benchmark under a specified metric.
\item \textbf{Independent chemical validation}: expert review, rule checks, quantum calculations, reaction predictors, or orthogonal tools support the claim.
\item \textbf{Prospective experimental success}: the proposed reaction or condition is executed successfully in the laboratory.
\item \textbf{Independent replication}: the result is reproduced across operators, equipment, laboratories, or independently constructed workflows.
\end{enumerate}
Level 3 evidence remains computational or benchmark-based and is not equivalent to experimental validation. Conversely, a prospective experiment can be highly informative even when its offline benchmark ranking is modest. Figure~\ref{fig:evidenceladder} visualizes the corresponding hierarchy and the increasing cost and realism associated with stronger evidence.

\subsection{Reproducible Reporting Requirements}
For reaction-prediction studies, reproducibility depends on the data-set version and split, preprocessing, stereochemical treatment, evaluation rule, uncertainty, and, where feasible, checks for training contamination. Commercial or API-served LLMs add another reproducibility requirement: the exact model identifier or snapshot date, prompts, and system instructions must be preserved because behavior can change while the product name remains constant.

Tool-augmented agents require an additional level of reporting. Ablations of tool access and orchestration logic are needed to determine whether an observed gain comes from the LLM, a chemistry tool, or the policy that connects them. ChemCrow and AutoLabs make these architectural components relatively explicit \cite{bran2024chemcrow,panapitiya2026autolabs}. Procedure-level benchmarks are also becoming more diagnostic. ChemReason-Bench, introduced at ACL 2026, contains 7,306 human-validated tasks derived from 500 organic reactions across ordering, step validation, condition validation, schema-constrained completion, contrastive choice, and evidence-grounded rationalization \cite{zhang2026chemreason}. It tests preservation of experimental procedure logic rather than only chemistry question answering or product-string regeneration.

For autonomous laboratories, the corresponding record includes the physical action space, safety interlocks, human intervention points, device failures, calibration, replicate experiments, unsuccessful trials, and the mechanism by which analytical outcomes become feedback. Without denominators and failure reports, successful demonstrations are vulnerable to survivorship bias.

\subsection{Chemically Meaningful Evaluation Metrics}
In parallel, chemistry-aware validation frameworks are beginning to formalize this distinction. A 2026 framework for benchmarking language-model synthesis planning separates high-level pathway coherence from the correctness and executability of experimental details \cite{zhang2026validation}. Although that study addresses materials synthesis, the methodological lesson transfers directly to organic synthesis: route-level plausibility, parameter-level correctness, and execution-level success are separate outcomes and cannot be collapsed into one language metric.

Top-$k$ exact match remains useful for some benchmark questions, but chemistry-aware measures are needed to characterize practical utility. Retrosynthesis evaluation benefits from separating recovery of the recorded route from chemically plausible alternatives and from reporting route-level feasibility, starting-material availability, stereochemical correctness, and diversity. Condition-optimization results are interpretable only when the experimental budget, baseline optimizer, uncertainty in yield or selectivity, and any safety or sustainability objectives are reported. Autonomous systems require task-completion, error-recovery, human-intervention, and reproducibility metrics in addition to natural-language correctness.

\section{Reliability, Interpretability, Safety, and Sustainability}
\subsection{Hallucination and Verification}
Despite these advances, LLMs can produce fluent but unsupported chemical statements. Reed showed in a molecular-property task that augmented generation and programmatically optimized prompting reduced TPSA prediction RMSE from 62.34 to 11.76 for the same underlying LLM \cite{reed2025hallucinations}. The result does not establish synthesis accuracy, but it demonstrates that inference-time scaffolding can materially change chemical reliability.

External chemistry tools can move calculations, database lookups, structure conversion, and reaction prediction out of free-form generation and into deterministic or specialist components \cite{bran2024chemcrow}. This does not eliminate hallucination; it changes the failure modes. An agent can select the wrong tool, misinterpret a valid result, or assemble an invalid sequence of individually correct operations. Validation is required at both the content level and the workflow level.

\subsection{Interpretability and Explanatory Fidelity}
Attention maps, natural-language rationales, and chain-of-thought-like outputs can be inspected, but they are not necessarily faithful explanations of model computation. A stronger synthesis-focused interpretability study would test whether reaction centers, functional-group constraints, or proposed intermediates causally influence predictions under controlled perturbations. Expert agreement, counterfactual consistency, and performance on withheld reaction classes provide stronger evidence than visual or verbal claims that a model ``understands'' a molecule.

In this respect, ChemBench provides relevant caution: leading LLMs can perform strongly on aggregate chemistry questions while showing weak performance on specific analytical, safety, preference, or structural reasoning tasks and overconfidence in incorrect outputs \cite{mirza2025chembench}. Reliability consequently depends on calibration and domain decomposition rather than a single general ``reasoning'' score.

\subsection{Safety, Access Control, and Dual Use}
Beyond reliability, chemical AI is inherently dual use. Earlier work showed that generative molecular-design systems developed for therapeutic objectives could be redirected toward toxic objectives \cite{urbina2022dual}. For LLMs, the risk includes unsafe procedural advice, bypass of access controls, synthesis guidance for regulated chemicals, and overconfident misinformation to non-expert users.

ChemSafetyBench, a 2024 preprint, contains more than 30,000 samples spanning chemical properties, legality of use, synthesis methods, and adversarial settings \cite{zhao2024chemsafety}. Its performance findings remain provisional because the benchmark has not yet passed peer review, but the resource illustrates a broader point: chemical safety requires domain-specific evaluation rather than generic chatbot safety scores.

Robust autonomous synthesis also requires a separation between capability and permission. In such an architecture, the language model proposes or interprets actions but does not hold unrestricted authority to convert arbitrary text into physical execution. Relevant controls include identity and access checks where appropriate, chemical-inventory restrictions, incompatibility screening, action allow-lists, hardware limits, audit logs, staged authorization, and an independent emergency-stop pathway. Safety therefore cannot depend solely on one generative model approving another.

\subsection{Sustainability as a Multi-Objective Evaluation Problem}
A related evaluation problem concerns sustainability. Algorithm-assisted synthesis is increasingly framed as a multi-objective problem in which yield, resource use, waste, route choice, and biological or chemical manufacturing options may need to be optimized jointly \cite{grzybowski2025sustainable}. This supports treating sustainability as an explicit vector of endpoints rather than as an inferred benefit of automation.

Terms such as ``green'' or ``sustainable'' synthesis require explicit quantitative endpoints. Higher yield can coexist with greater solvent hazard, energy demand, process mass intensity, or waste. A defensible sustainability claim identifies the quantity being optimized, such as E-factor, process mass intensity, atom economy, solvent hazard, energy use, catalyst burden, toxicity, or a defined multi-objective function. Without such an endpoint, the statement that an LLM improves sustainability is not falsifiable.

Moreover, the same principle applies to self-driving laboratories. Autonomous optimization can reduce the number of experiments, but computational cost, consumables, analytical overhead, solvent use, and failed runs must all be included before claiming environmental benefit. Multi-objective optimization consequently provides a more informative direction than yield-only optimization.

\section{Emerging Directions Toward 2027}
Broader 2026 surveys of LLMs across chemistry and biology increasingly emphasize domain adaptation, multimodality, tool use, and scientific validation rather than conversational fluency alone \cite{ashyrmamatov2026survey}. The synthesis literature shows the same shift. Progress is increasingly determined by how models are combined with specialist tools, uncertainty-aware decision layers, laboratory interfaces, and validation mechanisms.

This change is especially important because 2026 also brought several broad reviews of general-purpose chemical models and organic-chemistry-driven AI methodology \cite{alampara2026gpm,chawla2026organicai}. A new review can no longer establish novelty by cataloguing LLM applications alone. The more useful contribution lies in organizing evidence, separating system components, harmonizing evaluation, and identifying which claims survive prospective testing and replication.

The frontier now extends across several layers of the synthesis stack. ChemMLLM develops a unified interface across text, SMILES, and molecular images \cite{tan2026chemMLLM}; ReactionSeek applies LLMs to large-scale literature mining \cite{li2026reactionseek}; strategy-aware planning is evaluated against expert chemists rather than only historical route strings \cite{bran2026reasoning}; and BO-ICL and GOLLuM combine language-derived representations with explicit uncertainty in sequential optimization \cite{ramos2026boicl,rankovic2026gollum}. At the experimental layer, AutoLabs, RoboChem-Flex, chemputation, and Flex-Cat emphasize self-correction, modularity, executable procedure languages, mixed-variable optimization, and long-running autonomous operation \cite{panapitiya2026autolabs,pilon2026robochemflex,pagel2026chemputation,bennett2026flexcat}.

Peer-reviewed papers do not capture every emerging direction, so preprints are considered separately from journal evidence. QFANG, a 2025 preprint, explores structured experimental-procedure generation with chemistry-guided reasoning and verifiable rewards \cite{liu2025qfang}. RetroReasoner, released in March 2026, applies supervised and reinforcement learning to explicit retrosynthetic disconnection rationales, including a round-trip reaction reward \cite{ko2026retroreasoner}. C3LM, posted in August 2026, targets the one-to-many nature of single-step retrosynthesis through diverse Top-$K$ plausible outputs trained on a reported corpus of approximately 45.6 million verified reactions \cite{zagribelnyy2026c3lm}. These studies provide useful frontier signals, although their claims remain provisional pending peer review and independent reproduction.

Across these studies, the more important pattern is architectural. The broad label ``AI chemist'' is being decomposed into knowledge retrieval, representation, planning, uncertainty-aware experiment selection, protocol compilation, device orchestration, perception, analytical interpretation, and safety. Modular systems make these components replaceable and independently testable. They also improve reproducibility because model versions, tool configurations, optimizers, and hardware interfaces can be documented and varied without rebuilding the complete workflow.

Provenance is becoming part of this systems view. A correct synthesis suggestion without traceable sources may be less scientifically useful than a slightly less concise answer linked to the primary literature. Likewise, an autonomous run is difficult to reproduce without a record of prompts, model versions, device states, calibration, and raw analytical data. Recent self-driving-laboratory work accordingly treats end-to-end provenance and interoperability as infrastructure requirements rather than optional reporting details \cite{canty2026sdl}. The laboratory notebook of an LLM-enabled system must capture both chemical and computational state.

\subsection{Interoperable and Networked Chemical Agents}
A notable late-2026 development is the emergence of standardized connectivity between agents and scientific tools. ChemLint, published online in JCIM on 28 August 2026, implements a Model Context Protocol (MCP) server that connects compatible language-model clients to deterministic local cheminformatics and machine-learning tools \cite{vantilborg2026chemlint}. MCP does not improve reaction chemistry by itself. Its importance lies in providing a machine-readable interface through which an agent can call external capabilities without embedding each tool directly into a bespoke application. ChemLint also records operations in a project manifest, linking interoperability with provenance and reproducibility.

For organic-synthesis agents, this connectivity layer could separate rapidly changing language-model front ends from more stable specialist infrastructure, including reaction databases, molecular operations, retrosynthesis engines, optimizers, instrument APIs, and safety services. Such separation makes component substitution and controlled ablation more practical. It also turns portability into an empirical question: can the same agent policy operate across different tool servers and laboratory environments when the interface is standardized? The relevant outcomes include versioning, permissions, failure transparency, transfer cost, and provenance rather than a single successful tool call.

Laboratory infrastructure is changing in parallel. In July 2026, the U.S. National Science Foundation announced an inaugural investment in 20 teams to establish a national network of AI-enabled Programmable Cloud Laboratory nodes \cite{nsf2026pcl}. The program describes remotely accessible laboratories linked through computational networking and shared data and AI standards. Because this is an infrastructure initiative rather than a peer-reviewed chemical-performance study, it is treated here as a signal of research direction. Its architectural implication is nevertheless substantial: future scientific agents may dispatch validated workflows to networked experimental resources rather than operate only within one bespoke robotic laboratory.

MCP-style interfaces, uncertainty-aware optimization, provenance-complete execution, and programmable laboratory networks together suggest that near-term progress will depend less on language-model scale than on system integration. A useful synthesis agent will need standardized tools, explicit uncertainty, permissions, provenance, and recovery policies at each transition from computational planning to physical action.

Three developments summarize this transition. Strategy-aware planning is being compared with expert judgments rather than only exact-match data sets \cite{bran2026reasoning}; hybrid frameworks such as DeepRetro combine language-model flexibility with conventional retrosynthesis engines and expert feedback \cite{sathyanarayana2026deepretro}; and laboratory systems increasingly emphasize executable procedures and self-correction, as illustrated by chemputation and AutoLabs \cite{pagel2026chemputation,panapitiya2026autolabs}. RoboChem-Flex provides an important counterpoint by showing that modular engineering and Bayesian optimization can advance closed-loop chemistry without placing an LLM at the center of control \cite{pilon2026robochemflex}.

These developments shift the research agenda from whether an LLM can write a plausible synthesis toward the following system-level questions:
\begin{itemize}[leftmargin=*]
\item Which parts of synthesis benefit from broad language priors, and which should remain deterministic or chemistry-specific?
\item How much performance survives scaffold, reaction-class, laboratory, and model-version shift?
\item Can uncertainty estimates identify when a model should defer to tools or humans?
\item What minimum validation is required before a generated plan can trigger physical execution?
\item Can route quality be evaluated using experimentally meaningful multi-objective criteria rather than patent-route exact match alone?
\item Which autonomy components actually reduce labor or experimental budget after accounting for integration and recovery costs?
\item How should safety permissions scale with the evidentiary maturity of the system?
\item Can agent policies and verification logic transfer across standardized tool and programmable-laboratory interfaces without task-specific re-engineering?
\end{itemize}

\section{Reporting and Evaluation Priorities}
The literature reviewed above points to nine priorities for studies that combine language models with organic-synthesis prediction, planning, optimization, or laboratory execution. These priorities are not additional benchmarks in themselves; they define the information needed to determine which component of a system is responsible for an observed result and whether that result is likely to transfer beyond the original study.

\textbf{1) Unit of analysis.} Model, retrieval layer, agent, optimizer, and robotic platform are distinct experimental units. Component-level ablation is therefore central to separating language-model capability from system-level performance.

\textbf{2) Task-specific benchmarking.} Broad chemical question answering and reaction prediction probe different capabilities. ChemBench characterizes general chemical competence, whereas USPTO-, ORD-, and HTE-style data sets provide reaction-specific evidence \cite{mirza2025chembench,lowe2017uspto,sagawa2025reactiont5}.

\textbf{3) Failure reporting.} Autonomous chemistry requires complete denominators. Successful syntheses are difficult to interpret without the number and nature of failed procedures, recovery attempts, and human interventions.

\textbf{4) Retrospective and prospective evidence.} Recovery of a patent reaction is fundamentally different from proposing and executing a new reaction. Explicit separation of benchmark, expert-validation, prospective, and replicated evidence prevents these levels from being conflated.

\textbf{5) Model and prompt state.} Hosted models change over time. Reproducibility therefore depends on recording the model identifier, access date or snapshot, system instructions, prompts, decoding configuration, and agent state in supplementary materials or an auditable repository.

\textbf{6) Strong non-LLM baselines.} Reaction transformers, graph models, conventional CASP, Gaussian processes, Bayesian optimization, and deterministic controllers remain competitive for well-defined tasks. The added value of an LLM is most convincing when it appears as improved interface flexibility, transfer, semantic integration, recovery, or orchestration relative to these established baselines.

\textbf{7) Calibration and abstention.} A small gain in exact-match accuracy has limited value if it is accompanied by systematic overconfidence. In experimentally consequential settings, uncertainty estimates, abstention behavior, and escalation to tools or human supervision are part of system performance.

\textbf{8) Safety architecture.} Hazard databases, authorization, physical interlocks, inventory constraints, and audit trails operate most reliably as external system layers. This separation keeps safety enforcement independent of the generative model whose outputs are being checked.

\textbf{9) Interoperability and provenance.} Agent-to-tool interfaces, manifests, API versions, permissions, and laboratory endpoints determine whether a workflow can be reproduced across software and physical environments. MCP-style connectivity and programmable cloud laboratories make cross-system transfer an experimentally measurable property rather than a local implementation detail \cite{vantilborg2026chemlint,nsf2026pcl}.

\begin{table*}[t]
\caption{Minimum reporting checklist for LLM-enabled organic-synthesis studies.}
\label{tab:checklist}
\centering
\small
\renewcommand{\arraystretch}{1.0}
\begin{tabularx}{\textwidth}{P{2.6cm}YY}
\toprule
\textbf{Layer} & \textbf{Minimum information} & \textbf{Why it matters} \\
\midrule
Model & Exact model/checkpoint or API identifier; access date; parameter/update status where available; prompt/system instructions; decoding settings. & Proprietary and hosted models change over time; unreproducible prompts can dominate the result. \\
Chemical data & Dataset release, split, deduplication, atom mapping, stereochemistry, reagent handling, negative data, leakage checks. & Reaction benchmarks are highly sensitive to preprocessing and historical bias. \\
Tools/agent & Tool inventory, retrieval sources, tool-selection policy, memory/state, retry rules, verification steps, ablations. & Separates LLM behavior from external deterministic capabilities. \\
Optimization & Objective(s), surrogate model, uncertainty model, acquisition rule, experimental budget, baselines, stopping criteria. & Makes sample efficiency and claimed optimization gains interpretable. \\
Robotics & Hardware, action space, calibration, safety interlocks, human intervention points, device-error recovery, unsuccessful runs. & Physical autonomy can fail independently of language/planning quality. \\
Validation & Chemical-validity checks, expert review, prospective experiments, analytical characterization, replicates, independent reproduction where possible. & Distinguishes fluent output from reliable chemistry. \\
Provenance & Source references, raw data, agent logs, code/protocol versions, device metadata, timestamps. & Enables auditability and cross-laboratory transfer. \\
\bottomrule
\end{tabularx}
\end{table*}

\subsection{Testable Propositions for Future Work}
The preceding synthesis also supports six propositions that can be examined experimentally. Framing these ideas as comparative hypotheses helps move the field beyond demonstrations of possibility toward tests of which architectural choices produce reliable chemical performance.

\textbf{P1. Tool access and verification are expected to contribute more than naive model scaling on chemically constrained out-of-distribution tasks.} A factorial design that varies model scale and tool access can test whether larger models primarily improve fluent explanation or also improve chemically valid execution across reaction-family and temporal splits.

\textbf{P2. General chemistry question-answering performance is expected to be a weak predictor of prospective synthesis success.} ChemBench-like competence, reaction-prediction accuracy, route-quality assessment, and robotic task completion measure different capabilities. Their relationships can be quantified by reporting benchmark scores and blinded prospective execution for multiple systems under a common protocol.

\textbf{P3. Closed-loop performance is expected to depend more strongly on calibrated uncertainty and recovery policy than on one-shot generation accuracy.} An agent that recognizes uncertainty, retries safely, or requests human intervention may outperform a more accurate but overconfident generator. Holding the hardware and reaction family constant while varying calibration, abstention thresholds, and recovery rules provides a direct test.

\textbf{P4. Multimodal literature understanding is likely to constrain literature-to-protocol automation.} Errors in scheme recognition, reagent-role assignment, and cross-document grounding are expected to propagate into downstream protocol failures even when the planner and robot are held constant. RxnBench and ChemReason-Bench make this relationship experimentally tractable \cite{li2026rxnbench,zhang2026chemreason}.

\textbf{P5. Independent replication is likely to reorder apparent system rankings.} Systems evaluated on bespoke hardware in a single laboratory may lose relative advantage when protocols are transferred across operators, instruments, and inventories. Transfer cost, intervention rate, failed runs, and calibration burden therefore provide important complements to successful synthesis outcomes.

\textbf{P6. Standardized tool interfaces are expected to improve portability more strongly than single-site benchmark accuracy.} MCP-style connectivity and programmable laboratory interfaces are designed to decouple agents from bespoke integrations. Their principal benefit is therefore expected to appear as lower transfer cost, fewer interface-specific failures, and more reproducible provenance across heterogeneous environments, rather than as a large increase on one fixed benchmark \cite{vantilborg2026chemlint,nsf2026pcl}. This proposition can be tested by holding the planning policy constant while varying tool servers, API implementations, or laboratory nodes.

These propositions define a comparative research program in which model quality is evaluated together with the tools, uncertainty estimates, interfaces, and experimental conditions that determine real synthesis performance.

\section{Conclusion}
Large language models have become consequential components of organic-synthesis software and laboratory automation, although their scientific role is inseparable from the computational and physical systems in which they operate. Reaction prediction, chemical sequence learning, and neural-guided retrosynthesis were already mature research directions before the current LLM wave. Contemporary language models extend this foundation by providing natural-language interfaces, broad scientific priors, tool orchestration, code generation, and flexible coordination across heterogeneous tasks. These capabilities are most valuable when they complement, rather than obscure, the specialist models and deterministic infrastructure that supply chemical validity and experimental control.

The evidence available through 4 September 2026 shows several clear advances. Chemistry-adapted language models and reaction foundation models achieve strong performance on defined reaction tasks, as illustrated by Chemma and ReactionT5 \cite{zhang2025chemma,sagawa2025reactiont5}. Tool-augmented systems such as ChemCrow demonstrate how a general LLM can coordinate chemistry-specific software and information sources \cite{bran2024chemcrow}. At the experimental level, platforms including LLM-RDF, ChemAgents, and AutoLabs show that natural-language intent can be connected to increasingly sophisticated physical workflows \cite{ruan2024llmrdf,song2025chemagents,panapitiya2026autolabs}. In each case, however, the observed capability arises from a composite architecture that includes planning, optimization, robotics, sensing, validation, and, in many settings, human oversight.

The most important transition now occurring is from isolated model demonstrations to integrated scientific systems. Uncertainty-aware sequential optimization, executable procedure languages, self-correcting agents, standardized tool interfaces, provenance-aware execution, and programmable laboratory networks all point in this direction. Together, they shift attention from whether an LLM can generate a plausible synthesis to whether a complete human-AI-tool-laboratory system can make calibrated decisions, recover from failure, preserve a traceable computational and experimental state, and transfer across environments.

Progress toward this form of integrated chemical intelligence will depend less on a single benchmark score than on the consistency of evidence across levels. Historical reaction benchmarks remain valuable for model development, but prospective experiments, failure reporting, calibration under distribution shift, and independent replication are needed to establish scientific reliability. Organic synthesis therefore offers a particularly demanding test of agentic AI: useful systems must connect symbolic and statistical reasoning to irreversible physical actions while preserving chemical validity, safety, and reproducibility. Meeting these requirements will determine whether LLM-enabled synthesis develops into durable scientific infrastructure or remains a collection of impressive but difficult-to-transfer demonstrations.

\section*{Data and Software Availability}
No new experimental data set or software package was generated as a scientific result of this critical Review. The evidence base consists of the published sources cited in the manuscript. A claim-verification ledger, figure files, and the manuscript source are included in the accompanying submission package to support editorial and coauthor checking.

\balance

\end{document}